\documentclass[review]{elsarticle}

\usepackage{lineno,hyperref}

\journal{Journal of Information Fusion}
\usepackage{amsmath}
\usepackage{amsfonts}
\usepackage{array}
\usepackage{multirow}
\usepackage{booktabs}
\usepackage{url}

\usepackage{graphicx}
\usepackage{amssymb}
\usepackage{booktabs}
\usepackage{enumitem}
\usepackage{bbm}
\usepackage[ruled,linesnumbered]{algorithm2e}
\usepackage{color}
\usepackage[table]{xcolor}

\begin{document}

\begin{frontmatter}

\title{TransFuse: A Unified Transformer-based Image Fusion Framework using Self-supervised Learning}

\author{Linhao Qu\fnref{firstAuthor}}
\ead{lhqu20@fudan.edu.cn}
\author{Shaolei Liu\fnref{firstAuthor}}
\ead{slliu@fudan.edu.cn}
\author{Manning Wang}
\author{Shiman Li}
\author{Siqi Yin}
\author{Qin Qiao\corref{mycorrespondingauthor}}
\ead{qinqiao@fudan.edu.cn}
\author{Zhijian Song\corref{mycorrespondingauthor}}
\ead{zjsong@fudan.edu.cn}

\fntext[firstAuthor]{Linhao Qu and Shaolei Liu contributed equally to this work.}
\cortext[mycorrespondingauthor]{Corresponding author}

\address{Shanghai Key Lab of Medical Image Computing and Computer Assisted Intervention, Digital Medical Research Center, School of Basic Medical Science, Fudan University, Shanghai 200032, China}

\begin{abstract}
Image fusion is a technique to integrate information from multiple source images with complementary information to improve the richness of a single image. Due to insufficient task-specific training data and corresponding ground truth, most existing end-to-end image fusion methods easily fall into overfitting or tedious parameter optimization processes. Two-stage methods avoid the need of large amount of task-specific training data by training encoder-decoder network on large natural image datasets and utilizing the extracted features for fusion, but the domain gap between natural images and different fusion tasks results in limited performance. In this study, we design a novel encoder-decoder based image fusion framework and propose a destruction-reconstruction based self-supervised training scheme to encourage the network to learn task-specific features. Specifically, we propose three destruction-reconstruction self-supervised auxiliary tasks for multi-modal image fusion, multi-exposure image fusion and multi-focus image fusion based on pixel intensity non-linear transformation, brightness transformation and noise transformation, respectively. In order to encourage different fusion tasks to promote each other and increase the generalizability of the trained network, we integrate the three self-supervised auxiliary tasks by randomly choosing one of them to destroy a natural image in model training. In addition, we design a new encoder that combines CNN and Transformer for feature extraction, so that the trained model can exploit both local and global information.
Extensive experiments on multi-modal image fusion, multi-exposure image fusion and multi-focus image fusion tasks demonstrate that our proposed method achieves the state-of-the-art performance in both subjective and objective evaluations. The code will be publicly available soon.
\end{abstract}

\begin{keyword}
image fusion, Transformer, self-supervised learning, deep learning
\end{keyword}

\end{frontmatter}


\section{Introduction}
By integrating information from multiple source images with complementary information and fusing them into one fused image, image fusion technique can generate high-quality images and compensate for the inherent defects of a single imaging sensor \cite{1}. Image fusion has a wide range of applications \cite{1,2,3,4}. For example, in military applications, the fusion of infrared and visible images can be used for reconnaissance as well as night vision \cite{5,6,7,8}. In the medical field, fusing images of different modalities (e.g., Computed Tomography (CT) and Magnetic Resonance Imaging (MRI)) can assist the clinicians in clinical diagnosis and treatment \cite{9,10}. In the field of consumer electronics, multi-exposure image fusion can be employed to generate high dynamic range images for mobile devices \cite{11,12,13}, while multi-focus image fusion can be applied for refocusing algorithms \cite{14,15,16}.

Typically, image fusion tasks can be classified into multi-modal image fusion, multi-exposure image fusion and multi-focus image fusion. Although most studies focus on a certain fusion task, the design of unified image fusion frameworks that can be applied to different tasks is gradually becoming a significant research direction \cite{17,18,19,20}. The reason not only lies in the generality of a unified image fusion framework for multiple tasks, but also rests with the finding that these frameworks can achieve better performance when they are jointly trained on different tasks than being trained on a specific task \cite{17}. 

Existing image fusion methods can be classified into two categories: traditional methods \cite{1,3,21,22,23,24,25,26,27,28,29,30,31} and deep learning-based methods \cite{17,18,19,20,32,33,34,35,36,37}. Although traditional image fusion methods have achieved promising performance before the deep learning era, the hand-crafted feature extraction approaches limit further performance improvement. Moreover, these methods can only be used in a specific task due to their poor generalization capability. Deep learning-based image fusion methods have alleviated such limitation thanks to their powerful feature extraction capability, and gradually become the mainstream approaches. In these methods, the source images are fed into a deep neural network and the output of the network is the fused image. 

According to how the fusion network is trained, deep learning-based methods can be further divided into end-to-end methods \cite{17,18,19,20,32,34,35} and two-stage methods \cite{33,36,37}. In end-to-end methods, the fusion network is trained directly either in a supervised manner using synthetic ground truth fused images or in an unsupervised manner using loss functions defined on the similarity between the fused image and the input source images. However, the end-to-end methods require a large number of task-specific training images, which is difficult or expensive to collect in the image fusion field. Although some training datasets are constructed for specific fusion tasks \cite{17,75,76,77}, they are not comparable in size to large natural image datasets (e.g., ImageNet \cite{78}, COCO \cite{64}). Therefore, end-to-end image fusion methods easily fall into overfitting or tedious parameter optimization processes due to insufficient training data. In order to obtain more training data, most end-to-end methods \cite{17,19,20,32,34,35} divide the source images into small patches during the training process, which corrupts the semantic information of the whole image and also makes the network difficult to model the global features, leading to inferior fusion performance. Instead, the two-stage methods first train an encoder-decoder network on a large natural image dataset through image reconstruction. Then, the trained encoder is used to extract feature maps from the source images, and the feature maps are fused and further decoded to generate the fused image using the trained decoder. The advantage of two-stage methods is that the encoder and the decoder can be trained on large natural image datasets and can avoid the need of large amount of task-specific training data, which makes them more flexible and stable. 

However, the two-stage methods still have several unsolved issues that hinder further performance improvement. First, the current two-stage methods \cite{33,36,37} purely focus on the reconstruction of natural images but the domain gap between natural images and fusion tasks results in poor generalization of the extracted features and inferior performance. In addition, the same natural image dataset is usually used for different fusion tasks without considering task-specific features. Therefore, it is a prominent issue to enable the encoder-decoder network to be trained on large natural image datasets while learn task-specific image features at the same time. Second, recent studies have indicated \cite{17,18,19,20} that joint training on different tasks can help improve the performance on each single task. Therefore, how to design a joint training scheme for two-stage frameworks is another important issue. Third, all existing deep-learning-based image fusion methods utilize CNN for feature extraction, but it is difficult for CNN to model long-range dependencies due to its small receptive field \cite{38}. 

To address the issues mentioned above, in this paper, we propose a novel two-stage image fusion framework based on a new encoder-decoder network, which is named TransFuse. First, to enable our network to be trained on large natural image datasets and learn task-specific features at the same time, we design three destruction-reconstruction self-supervised auxiliary tasks for each of the three image fusion tasks, multi-modal image fusion, multi-exposure image fusion and multi-focus image fusion. Instead of simply inputting a natural image into the encoder and using the decoder to reconstruct it, we destroy the natural image before inputting it into the encoder and design a specific way of destruction for each fusion task. By enforcing the encoder-decoder network to reconstruct a destroyed image, we can make the network to learn better task-specific image features. Second, in order to encourage different fusion tasks to promote each other and increase the generalizability of the trained network, we integrate the three self-supervised auxiliary tasks by randomly choosing one of them to destroy a natural image in model training. Third, to compensate for the defects of CNN in modeling long-range dependencies, we design a new encoder that combines CNN and Transformer to exploit both local and global information in feature extraction. We conduct extensive experiments to demonstrate the effectiveness of each component of our framework.

The main contributions are summarized as follows.
\begin{itemize}
\item We design a novel encoder-decoder based image fusion framework and propose a destruction-reconstruction based self-supervised training scheme to encourage the network to learn task-specific features.
\item We propose to use three transformations to destroy a natural image, pixel intensity non-linear transformation, brightness transformation, and noise transformation for multi-modal image fusion, multi-exposure image fusion and multi-focus image fusion, respectively. We integrate the three transformations by randomly selecting one of them to destroy an image in model training so as to make the trained model extract more generalizable features and achieve higher performance on each task.
\item We design a new encoder that combines CNN and Transformer for feature extraction, so that the trained model can exploit both local and global information.
\item Extensive experiments on multi-modal image fusion, multi-exposure image fusion and multi-focus image fusion tasks showed that our proposed method achieved new state-of-the-art performance in both subjective and objective evaluations.
\end{itemize}

\section{Related Work}
In this section, we will briefly review the most representative methods in the field of image fusion in recent years, including traditional methods and deep learning-based methods. The latter ones are divided into end-to-end methods and two-stage methods. Subsequently, we will introduce the related works of self-supervised learning and Transformer in the field of computer vision and their application potential in image fusion field.
\subsection{Image Fusion}
\subsubsection{Traditional Image Fusion Algorithms}
Traditional image fusion methods can be classified into spatial domain-based methods, transform domain-based methods, sparse representation and dictionary learning-based methods. Spatial domain-based methods usually calculate a weighted average of local or pixel-level saliency of two source images \cite{3,21,22,23} to obtain the fused image.

Transform domain-based methods firstly transform source images into a transform domain (e.g., wavelet domain) to obtain different frequency components. Then the corresponding components are fused by appropriate fusion rules, and finally the fused image is obtained by inverse transform. Commonly used transforms include Laplace pyramid (LP) \cite{24}, low-pass pyramid (RP) \cite{25}, discrete wavelet (DWT) \cite{1}, discrete cosine (DCT) \cite{26}, curvelet transform (CVT) \cite{27} and shearlet transform (Shearlet) \cite{28}, etc.

New methods based on sparse representation and dictionary learning have also emerged in recent years. For example, Liu \cite{29} et al. proposed JSR and JSRSD based on joint sparse representation and saliency detection. They first obtained the global and local representative maps of the source images based on sparse coefficients, and then combined them by a representative detection model to generate the overall representative map. Finally, a weighted fusion algorithm was adopted to obtain the fused image based on the overall representative map.

Although aforementioned methods have achieved good results, their performance are still limited due to the following two aspects. First, the complicated manually designed feature extraction approaches usually fail to effectively preserve important information in the source images and cause artifacts in the fused image. Second, the feature extraction methods are usually designed for a specific task so it is difficult to adapt them to other tasks.
\subsubsection{Deep Learning-based Image Fusion Algorithms}
Due to the powerful feature extraction capability of CNN, deep learning-based methods have gradually become the mainstream approaches in the field of image fusion. Deep learning-based methods can be further divided into end-to-end methods and two-stage methods.
\paragraph{End-to-End Image Fusion Scheme} In end-to-end methods, the fusion network is directly trained either in a supervised manner using synthetic ground truth fusion images or in an unsupervised manner using loss functions defined on the similarity between the source images and the fused images.

Zhang et al \cite{18} proposed IFCNN, a supervised unified image fusion framework. They constructed a large multi-focus image fusion dataset with ground truth images and utilized a perceptual loss function for supervised training. Then the trained model was transferred to other fusion tasks. Prabhakar et al \cite{32} proposed DeepFuse, an unsupervised multi-exposure image fusion framework utilizing a no-reference quality metric as loss function. They designed a novel CNN-based network trained to learn the fusion operation without ground truth fusion images. Ma et al. \cite{34} proposed FusionGAN, an unsupervised infrared and visible image fusion framework based on GAN. The generator network is used to generate the fused image and the discriminator network makes sure that the texture details in visible images together with thermal radiation information in infrared images are retained in the fused image. On this basis, they further proposed DDcGAN \cite{35}, which can enhance the edges and the saliency of thermal targets by introducing a target-enhanced loss function and a dual discriminator structure. Zhang \cite{19} et al. proposed PMGI, an unsupervised unified image fusion network. They unified the image fusion problem into the proportional maintenance of texture and intensity information of the source images, and proposed a new loss function based on the gradient information and intensity information between the fused image and the source images for unsupervised training. Xu et al. \cite{17,20} proposed U2Fusion, an unsupervised unified image fusion network. They utilized a novel loss function based on adaptive information preservation degree for unsupervised training. During training, a pre-trained neural network was used to extract the features of the source images and these features are further used to calculate the adaptive information preservation degree. 

The above end-to-end models have achieved promising fusion performance, but both supervised and unsupervised methods require a large number of task-specific training images. Although several studies have constructed some training datasets for specific fusion tasks (e.g., RoadScene \cite{17} and TNO \cite{77} for infrared and visible image fusion; Harvard \cite{82} for medical image fusion; SICE \cite{75} for multi-exposure image fusion; and Lytro \cite{76} for multi-focus image fusion), their sizes are not comparable to that of large natural image datasets (e.g., ImageNet \cite{78}, COCO \cite{64}). Insufficient training data tends to cause overfitting or complex parameter optimization. Besides, in order to obtain more training data, most end-to-end methods \cite{17,19,20,32,34,35} divide the source images into a large number of small patches during the training process, which corrupts the semantic information of the whole image and also prevents the network from modeling global features of the image. Above limitations hinder further performance improvement of the end-to-end methods.

\paragraph{Two-stage Image Fusion Scheme} In two-stage methods, an encoder-decoder network is first trained on large natural image datasets by image reconstruction in the first stage. In the second stage, the trained encoder is used to extract feature maps from source images and the features maps are then fused and decoded by the trained decoder to generate the fused image. 

Li et al \cite{33} presented the first two-stage method DenseFuse, for infrared and visible image fusion, and in the fusion process, they introduced l1-Norm based fusion rule to fuse the feature maps. Ma et al. \cite{36} proposed SESF-Fuse, a new two-stage framework for multi-focus image fusion. They utilized spatial frequency to measure activity level of the feature maps and acquired decision maps based on the activity level. Finally, the decision maps were used to obtain the fusion results. Liu et al \cite{37} proposed WaveFuse, a unified image fusion framework based on multi-scale discrete wavelet transform (DWT). In the fusion stage, the feature maps of source images are transformed into several components in the wavelet domain and fusion was done component by component. Then, the fused feature maps were reconstructed using inverse DWT, which was further inputted into the trained decoder to generate the fused image.

The advantage of two-stage methods is that the training can be performed self-supervisedly on accessible large natural image datasets without the need of scene-specific datasets or ground truth fusion images. However, the current two-stage frameworks simply perform natural image reconstruction and cannot enable the encoder-decoder network to learn task-specific image features. In this paper, we propose a novel two-stage image fusion framework which retains the above advantages of two-stage methods. At the same time, we propose to use three destruction-reconstruction auxiliary tasks to encourage the network to learn task-specific image features.
\subsection{Self-supervised Learning}
Self-supervised learning is a branch of unsupervised learning, which automatically generate its own supervised labels from large-scale unsupervised data. A network can learn valuable representations for downstream tasks when it is trained to perform auxiliary tasks using the generated labels \cite{39}. Self-supervised learning has been successfully applied in many fields such as computer vision and natural language processing \cite{39}. In the field of computer vision, self-supervised auxiliary tasks include angle prediction \cite{40}, image puzzles \cite{41}, image coloring \cite{42} and image reconstruction \cite{43}. Extensive studies \cite{44,45,46,47} have shown that by designing suitable auxiliary tasks, self-supervised learning can learn effective and generalized image feature representations. In this study, we design three self-supervised reconstruction auxiliary tasks for three image fusion tasks and we further integrate the three auxiliary tasks using random combination to further improve the performance of the trained model. 
\subsection{Transformer}
In the field of computer vision, CNNs and its variants have been widely used due to its powerful feature extraction capability. Nevertheless, CNN fail to establish long-range dependencies because of its inherent small receptive field. Unfortunately, almost all existing image fusion architectures are based on CNN, and the global information are not fully exploited. In the field of natural language processing, Transformer has achieved satisfying results in modeling global dependencies through the self-attention mechanism. A great number of works have emerged to utilize Transformer as an alternative to CNN in the field of computer vision and have achieved fantastic results in different tasks, such as image classification \cite{38,48}, object detection \cite{49,50,51,52,53}, image segmentation \cite{54,55} and image generation \cite{56} etc. Extensive studies \cite{17,18,32,33,34,35,36,37} have shown that the ability to extract effective image feature representation is the key to improve image fusion performance. To remedy the deficiency of establishing long-range dependencies within current CNN-based image fusion architectures, we design a new feature extraction module that combines CNN with Transformer to allow the network to exploit both local and global information more comprehensively.

\section{Method}
\subsection{Framework Overview}
\begin{figure*}[htbp]
    \centering
    \includegraphics[scale=0.20]{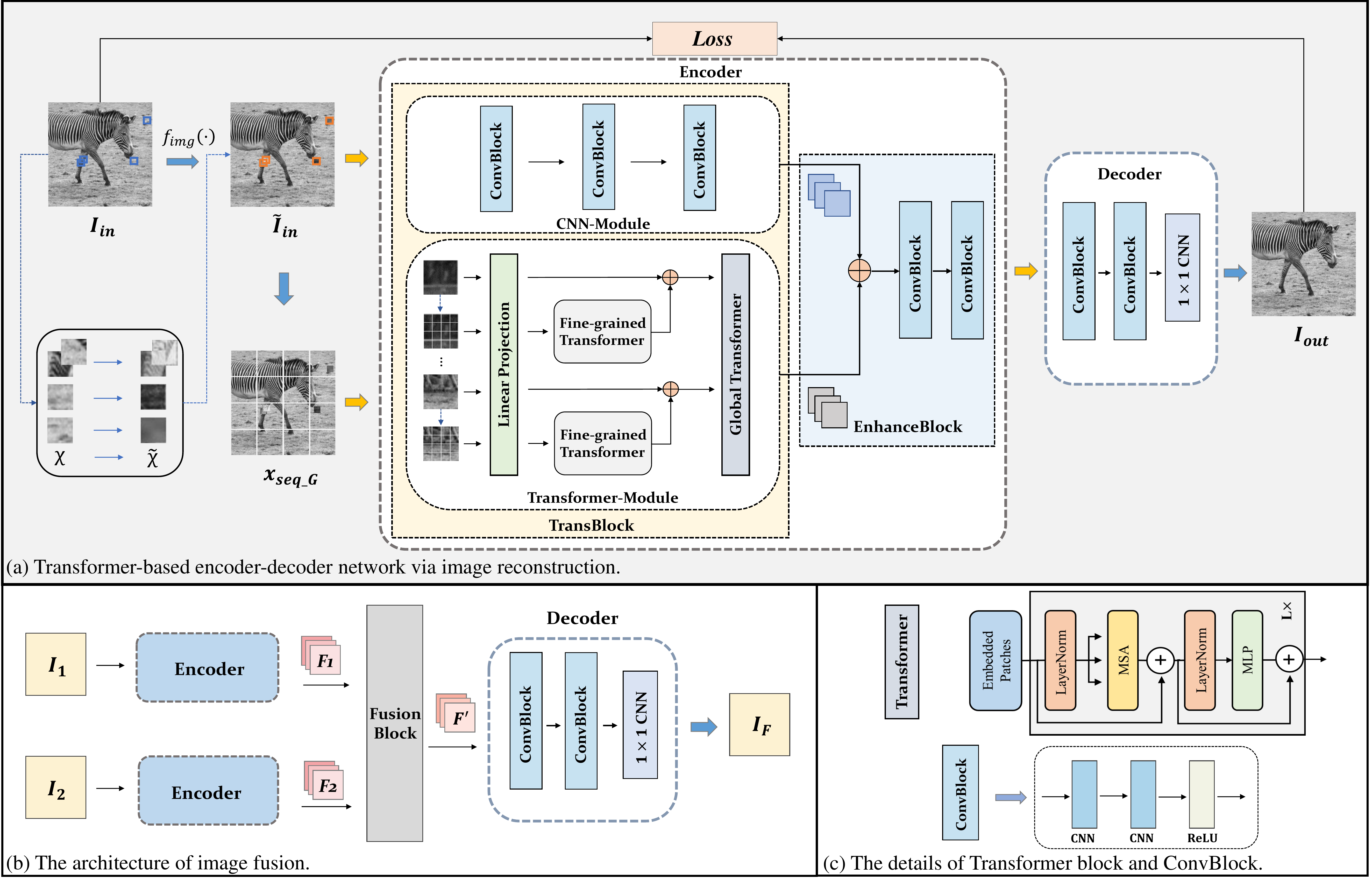}
    \caption{The two-stage image fusion framework based on an encoder-decoder network. (a) The framework of training a Transformer-based encoder-decoder network via image reconstruction. (b) The architecture of fusing two source images using the trained encoder and decoder. (c) Details of the Transformer block and ConvBlock.}
    \label{figure1}
\end{figure*}

The overall architecture of our framework is shown in Fig. \ref{figure1}. We train the encoder-decoder network by image reconstruction on a large natural image dataset. Different from existing two-stage methods, which directly input the original natural images into the encoder-decoder network for reconstruction, we first destroy the original images before input them into the encoder. Concretely, given an original training image ${\ I}_{in\ }\in\mathbb{R}^{H\times W}$, we first randomly generate $n$ image subregions $x_i\in\mathbb{R}^{H_{sub}\times W_{sub}} \left(i=1,2,\ldots n\right)$ to form a set $\chi$. For each subregion $x_i$ in $\chi$, we randomly apply pixel intensity non-linear transformation, brightness transformation and noise transformation to obtain a set of transformed subregions $\widetilde{\chi}$ and the destroyed input image ${\widetilde{I}}_{in\ }$. Then, we input the destroyed image ${\widetilde{I}}_{in\ }$ into the encoder, which consists of a feature extraction module TransBlock and a feature enhancement module EnhanceBlock. Finally, the extracted features are input to the decoder to reconstruct the original training image. The detailed structures of the encoder and the decoder are described in Section 3.2, and the three different task-specific transformations are introduced in detail in Section 3.3.

As shown in Fig. \ref{figure1} (b), the fusion framework consists of two parameter-shared encoders, a fusion block and a decoder. Note that both the encoder and the decoder are trained in the first stage, and the fusion block has no parameters to be trained, which ensures the simplicity and efficiency of the fusion framework. Specifically, two source images $I_k\ (k=1,2)$, are first input to the encoder for feature encoding, then the extracted feature maps $F_1$, $F_2$ are fused by the fusion block to obtain the fused feature maps $F^\prime$. Finally, $F^\prime$ is decoded by the decoder to reconstruct the fused image $I_F$. Since the important information differs in the fused images of different tasks, different fusion rules are used for the characteristics of different tasks. The fusion rules are introduced in detail in Section 3.4.
\subsection{Transformer-Based Encoder-Decoder Framework for Training}
\subsubsection{Encoder-Decoder Network via Image Reconstruction}
We train an encoder-decoder network via image reconstruction for image fusion, and the network architecture is shown in Fig. \ref{figure1} (a). Given a training image ${\ I}_{in\ }\in\mathbb{R}^{H\times W}$, we randomly generate $n$ image subregions $x_i\in\mathbb{R}^{H_{sub}\times W_{sub}} \left(i=1,2,\ldots n\right)$ to form a set $\chi$ for transformation. For each subregion $x_i$\ in $\chi$ to be transformed, we randomly apply three transformations (the three transformations for different fusion tasks are detailed in Section 3.3) to obtain the set of transformed subregions $\widetilde{\chi}$ and the transformed input image${\ \widetilde{I}}_{in\ }$.
\begin{equation}
    \widetilde{\chi}=f\left(\chi\right) \label{eq1}
\end{equation}
\begin{equation}
    {\widetilde{I}}_{in\ }=f_{img}\left(I_{in}\right) \label{eq2}
\end{equation}
where $\widetilde{\chi}=\{{\widetilde{x}}_1,{\widetilde{x}}_2,\ldots,{\widetilde{x}}_n\}$, $f(\cdot)$ denotes the transform function for a subregion set, ${\widetilde{x}}_{i\ }$ is the transformed image subregion; ${\ \widetilde{I}}_{in\ }$is the transformed input image, and $f_{img}(\cdot)$ refers to the image transform function. 

We perform a self-supervised image reconstruction task for model training so that the encoder-decoder network learns an inverse mapping $g(\cdot)$ that reconstructs the original image from the transformed input image.
\begin{equation}
    g\left({\widetilde{I}}_{in}\right)=\ I_{in}=f_{img}^{-1}({\widetilde{I}}_{in}) \label{eq3}
\end{equation}

Notice that we do not transform the whole input image, but a random selection of subregions of it. The reconstruction of the image is conducted at the whole image level .

As shown in Fig. \ref{figure1} (a), the encoder contains a feature extraction module named TransBlock, and a feature enhancement module called EnhanceBlock. TransBlock contains two sub-modules, the CNN-Module and the Transformer-Module based on CNN and Transformer, respectively. The specific design of TransBlock is described in Section 3.2.2. The EnhanceBlock further aggregates and enhances the feature maps of the CNN-Module and the Transformer-Module in TransBlock, thus allowing the encoder to better integrate local and global features. In particular, we concatenate the encoded features from the CNN-Module and Transformer-Module, and then input them into two ConvBlock layers to achieve the integration and feature enhancement. As shown in Fig. \ref{figure1} (c), each ConvBlock consists of two convolutional layers with kernel size of 3$\times$3, padding of 1 and a ReLU activation layer. The decoder contains two sequentially connected layers of ConvBlock followed by a 1$\times$1 convolution to reconstruct the original image.

\subsubsection{TransBlock: A Powerful Local and Global Feature Extractor}
Inspired by ViT \cite{38} and TNT \cite{79}, we combine CNN and Transformer to propose a powerful feature extraction module, TransBlock, to exploit both local and global information in source images. As shown in Fig. \ref{figure1} (a), the TransBlock consists of two feature extraction sub-modules, the CNN-Module and the Transformer-Module. CNN-Module consists of three sequentially connected ConvBlocks. Transformer-Module are designed with a Fine-grained Transformer for local features modeling and a Global Transformer for global features modeling. Concretely, we first divide the transformed input image ${\ \widetilde{I}}_{in\ }\in\mathbb{R}^{H\times W}$ into $N_G$ patches with the size of $P_G$ $\times$ $P_G$ and construct a global sequence of images $x_{seq_G}\in\mathbb{R}^{N_G\times{P_G}^2}$, where $x_{seq_G}=\left\{{x_G}^{j_G}\right\} \left(j_G=1,2,\ldots N_G\right)$, $N_G=HW/{P_G}^2$, $(P_G\times P_G)$ is the size of the divided patches. To further capture finer-grained features, we further divide each patches in the global sequence $x_{seq_G}$ into smaller sub-patches and construct the local sequence $x_{seq_L}\in\mathbb{R}^{N_L\times{P_L}^2}$, where $x_{seq_L}=\left\{{x_L}^{j_L}\right\} \left(j_L=1,2,\ldots N_L\right)$, $N_L=HW/{P_G}^2{P_L}^2$, $(P_L\times P_L)$ is the size of divided sub-patches.

For the processing of local sequences, we use the Fine-grained Transformer with shared weights to learn fine-grained relative dependencies in images.

First, we input the local sequence $x_{seq_L}$\ into the Linear Projection layer to perform a linear mapping to obtain the encoded feature sequence.
\begin{equation}
    \varphi^{j_G}=\left[\varphi^{j_G,1},\ \varphi^{j_G,2},\cdots,\varphi^{j_G,j_L}\right],\ \varphi^{j_G,\ j_L}=Vec{\left({x_L}^{j_G,\ j_L}\right)W}+b \label{eq4}
\end{equation}
where ${x_L}^{j_G,\ j_L}$ refers to the $j_Lth$ sub-patch of the $j_Gth$ patch; $\varphi^{j_G,\ j_L}$\ represents the encoded features of the sub-patch; $Vec(\cdot)$ flattens the input patch to one-dimensional vector; $W$, $b$ mean the weights and bias of the Linear Projection layer, respectively.

We then input the encoded features of the local sequences into the Fine-grained Transformer. The Fine-grained Transformer adopts a standard Transformer structure similar to ViT \cite{38}, as shown in Fig. 1 (c).
\begin{align}
    \varphi_l^{\prime j_G} & =\varphi_{l-1}^{j_G}+MSA\left(LN\left(\varphi_{l-1}^{j_G}\right)\right) \notag \\
    \varphi_l^{j_G} & =\varphi_l^{\prime j_G}+MLP\left(LN\left(\varphi_l^{\prime j_G}\right)\right)
    \label{eq5}
\end{align}
where $l = 1,2\ldots L$ denotes the $lth$ block, and $L$ refers to the number of transformer blocks.

For the global sequence, similar to the local sequence, we first input the global sequence\ $x_{seq_G}$ into the Linear Projection layer. Afterwards, we linearly map the output of the corresponding Fine-grained Transformer of the local sequence and concatenate them to the encoded features of the global sequence.
\begin{align}
    \psi^{j_G} & =Vec{\left({x_G}^{j_G}\right)W}+b \notag \\ 
    \psi_{l-1}^{j_G} & =\psi_{l-1}^{j_G}+Vec\left(\varphi_{l-1}^{j_G}\right)W_{l-1}+b_{l-1}
    \label{eq6}
\end{align}
where $Vec(\cdot)$ flattens the input patch to one-dimensional vector; $W_{l-1}$,${\ b}_{l-1}$ stands for the weights and bias respectively.

Then, we input the interactive encoding of global and local sequences into the Global Transformer. The structure of the Global Transformer utilizes a standard Transformer structure similar to ViT \cite{38}, as shown in Fig. \ref{figure1} (c).
\begin{align}
    \psi_l^{\prime j_G} & =\psi_{l-1}^{j_G}+MSA\left(LN\left(\psi_{l-1}^{j_G}\right)\right) \notag \\
    \psi_l^{j_G} & =\psi_l^{\prime j_G}+MLP\left(LN\left(\psi_l^{\prime j_G}\right)\right)
    \label{eq7}
\end{align}

In general, the Fine-grained Transformer is designed to model fine-grained relative dependencies within an image patch, and thus extract local semantic features, while the Global Transformer is designed to model global relative dependencies within an image, and thus extract global features. 

\subsubsection{Loss Function}
We expect our network to learn more than just pixel-level image reconstruction, but also sufficiently capture the structural and gradient information in the images. Therefore, we use a loss function with three components,
\begin{equation}
    L = L_{mse} + \lambda_1L_{ssim} + \lambda_2L_{TV}
    \label{eq8}
\end{equation}
where $L_{mse}$\ is the Mean Square Error (MSE) loss function, $L_{ssim}$ denotes the Structural Similarity (SSIM) loss function \cite{82}, $L_{TV}$ represents the Total Variation (TV) loss function \cite{58}, and $\lambda_1$ and $\lambda_2$ are two coefficients used to balance each loss function, which are empirically set to 20 in the experiment.

The MSE loss is used for pixel-level reconstruction of images and it is defined as
\begin{equation}
    L_{mse}=||I_{out}-I_{in}{||}_2
    \label{eq9}
\end{equation}
where $I_{out}$ denotes the output image of the decoder and $I_{in}$ represents the input image.

The SSIM loss is used to make the network learn structural information of the images, and it is defined as
\begin{equation}
    L_{ssim}=1-SSIM(I_{out},I_{in})
    \label{eq10}
\end{equation}
\begin{equation}
    SSIM\left(X,Y\right)=\frac{\left(2\mu_X\mu_Y+C_1\right)\left(2\sigma_{XY}+C_2\right)}{\left({\mu_X}^2+{\mu_Y}^2+C_1\right)\left({\sigma_X}^2+{\sigma_Y}^2+C_2\right)}
    \label{eq11}
\end{equation}
where $\mu$ and $\sigma$ denote the mean and the standard deviation, respectively, and $\sigma_{XY}$ is the correlation between $X$ and $Y$. The C1 and C2 are two very small constants, empirically set to 0.02 and 0.06. The standard deviation of the Gaussian window is empirically set to 1.5.

The TV loss is used to preserve the gradient information in the images and further eliminate the noise during image reconstruction, and it is defined as follows,
\begin{equation}
    R\left(i,j\right)=I_{out}\left(i,j\right)-I_{in}\left(i,j\right)
    \label{eq12}
\end{equation}
\begin{equation}
    L_{TV}=\sum_{i,j}{(||R\left(i,j+1\right)-R(i,j){||}_2+||R\left(i+1,j\right)-R(i,j){||}_2)}
    \label{eq13}
\end{equation}
where $R$ is the difference between the original image and the reconstructed image, $||\ {||}_2$ is the $l_2$ norm, and $i$, $j$ represent the horizontal and vertical coordinates of the image pixels, respectively.
\begin{figure*}[htbp]
    \centering
    \includegraphics[scale=0.5]{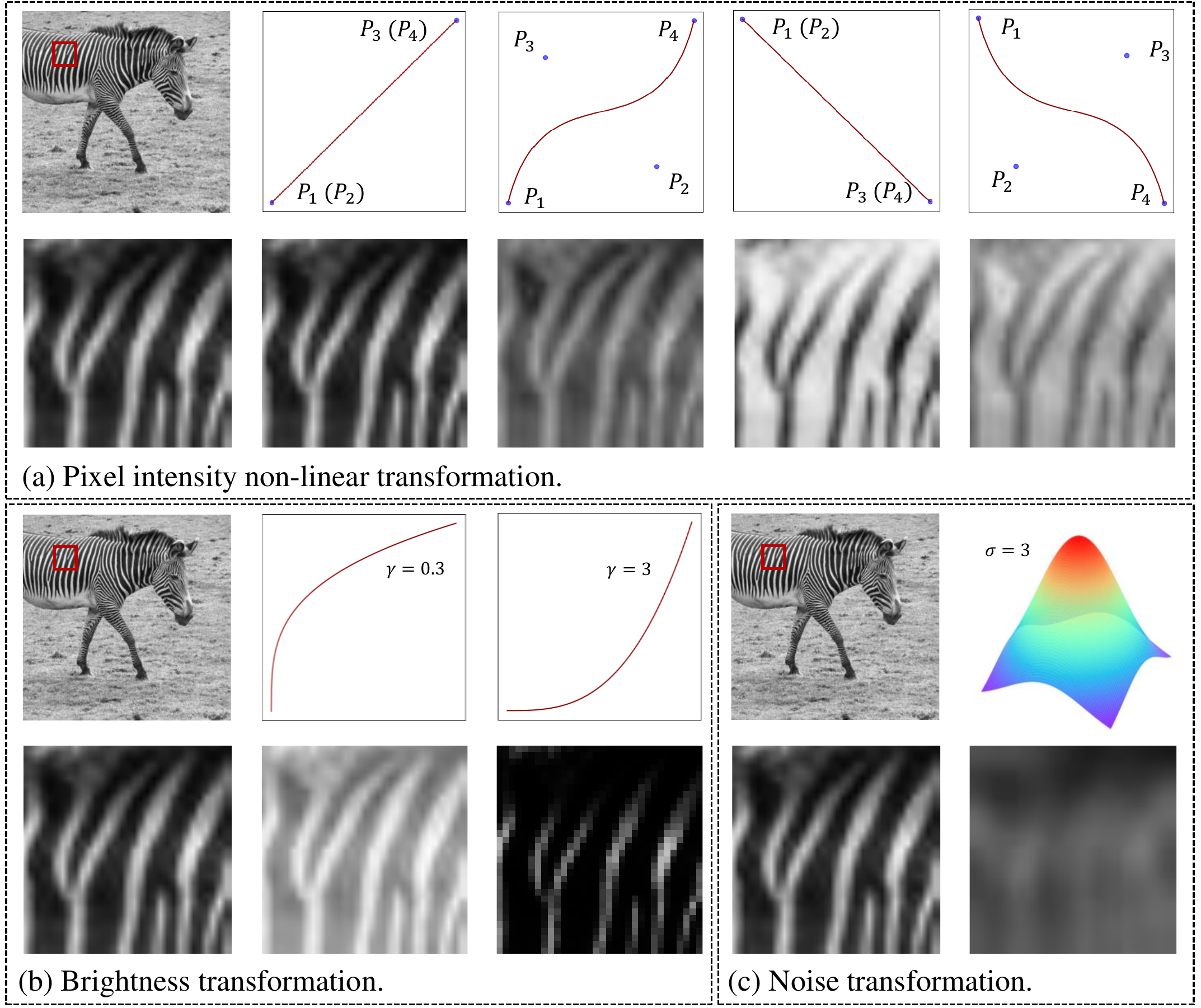}
    \caption{The three image transformations used to destroy a source image in destruction-reconstruction self-supervised auxiliary tasks. (a) Pixel intensity non-linear transformation for multi-modal image fusion. The first column shows the original image and one of its subregions and the second to fifth column show the image subregions after different non-linear transformations and corresponding Bessel transform curves. (b) Brightness transformation multi-exposure image fusion. The first column shows the original image and one of its subregions and the second to third column show the image subregions after Gamma transform and corresponding Gamma transform curves. (c) Noise transformation multi-focus image fusion. The first column shows the original image and one of its subregions and the second column shows the image subregions after the Gaussian blur transform and corresponding Gaussian function.}
    \label{figure2}
\end{figure*}

\subsection{Task-Specific Self-Supervised Training Scheme}
In this section, we introduce the three destruction-reconstruction auxiliary tasks designed for different image fusion tasks. For multi-modal, multi-exposure and multi-focus image fusion, the auxiliary task is based on pixel intensity non-linear transformation, brightness transformation and noise transformation, respectively. Specifically, for each subregion $x_i$\ in the set $\chi$ to be transformed, we randomly apply the three transforms to obtain the set of transformed subregions $\widetilde{\chi}$ and then the transformed input image ${\ \widetilde{I}}_{in\ }$. The transformed input image ${\ \widetilde{I}}_{in\ }$\ is input into the encoder-decoder network for reconstruction. This kind of destruction-reconstruction based auxiliary task enables our network to be trained on large natural image datasets and learn task-specific features at the same time.

\subsubsection{Pixel Intensity Non-Linear Transformation for Multi-Modal Image Fusion}
We design a novel self-supervised auxiliary task based on pixel intensity non-linear transformation for multi-modal image fusion. In multi-modal image fusion, different source images contain different kinds of information of modalities, and we hope the most important information from each modality can be retained in the fused image. Following \cite{17}, we study two scenarios of multi-modal image fusion, infrared and visible image fusion and multi-modal medical image fusion. In the former scenario, the most significant information is the thermal radiation in infrared images and the structural semantic information in visible images \cite{17,34}. In the later scenario, the most significant information is the functional response and the structural anatomical information in medical images \cite{59,60}. The above important fusion information is all reflected in the form of image pixel intensity distribution in the source images \cite{17}. Therefore, we propose a pixel intensity non-linear transformation to destroy the pixel intensity distribution first in the source images and then train the network to reconstruct the original pixel intensity. By doing this, our network can learn the pixel intensity information in the source images effectively.

Concretely, we use a smooth monotonic third-order Bézier Curve \cite{61} composed of four control points to implement the non-linear transformation. The four control points include two endpoints ($P_1$\ and $P_2$) and two midpoints ($P_3$ and $P_4$), defined as:
\begin{equation}
    P\ =\ {(1-t)}^3P_1\ +\ 3{(1-t)}^2tP_2\ +3\left(1-t\right)t^2P_3\ +\ t^3P_4\ \ \ t\in[0,1]
    \label{eq14}
\end{equation}
\begin{equation}
    {\widetilde{x}}_i\ =\ f_{{NL}}\left(x_i\right)=interp(x_i,P)
    \label{eq15}
\end{equation}
where, $t$ is a fractional value along the length of the line and interp indicates the interpolation function. Fig. \ref{figure2} (a) illustrate an original image subregion and the image subregions transformed by different Bessel transform curves. Specifically, we set two endpoints ${\ P}_1=(0,0)$ and ${\ P}_4=(1,1)$ to get a monotonically increasing Bessel transform curve. Then we randomly flip the curve to get a monotonically decreasing curve. And the midpoints ($P_3$ and $P_4$) are generated randomly for more variances. As shown in columns 2 and 4, when the midpoints $P_2$ and $P_3$ coincide with the two endpoints respectively, the transformation function is linear. The midpoints in columns 3 and 5 are randomly generated for more variances. Please note that by using a transformation curve like in column 4 and 5, the pixel intensity can be reversed.

\subsubsection{Brightness Transformation for Multi-Exposure Image Fusion}
For multi-exposure image fusion, we propose a self-supervised auxiliary task based on brightness transformation, encouraging the network to learn content and structural information at different exposure levels. In general, over-exposed images (images captured with long exposure time) have better content and structural information in dark regions, while under-exposed images (images captured with short exposure time) have better information in bright regions. Therefore, for multi-exposure image fusion, it is crucial to maintain appropriate luminance in the fused image while preserving abundant information \cite{17,32}. In this study, we design a brightness transformation to destroy the luminance of the source images and train the encoder-decoder network to reconstruct the original image. In this process, our network can learn well about the content and structural information of the images at different exposure levels, and thus can learn important fusion information for multi-exposure images.

The brightness transformation is implemented using the Gamma transform, a specific non-linear operation which is widely used to encode and decode brightness or trichromatic values in image and video processing \cite{62}. The Gamma transform is defined as
\begin{equation}
    \widetilde{\psi}\ =\ \Gamma(\psi)=\psi^{gamma}
    \label{eq16}
\end{equation}
where $\widetilde{\psi}$ and $\psi$ are the transformed pixel values and the original pixel values, respectively. Fig. \ref{figure2} (b) shows an original image subregion and the image subregions after brightness transformations with two different Gamma transformation curves. For each pixel in the selected subregion $x_i$, $gamma$ is empirically set as 0.3 to compress the brightness or 3 to enlarge the brightness.

\subsubsection{Noise Transformation for Multi-Focus Image Fusion}
For multi-focus image fusion, we propose a self-supervised auxiliary task based on noise transformation to enable the network to learn the variations of different depths of field (DoF) and maintain clear detail information. Due to the limitation of a camera’s DoF, it is very difficult to obtain an all-in-focus image within one shot. Objects within the DoF can maintain clear detail information, but the scene content outside the DoF is blurry. The main objective of multi-focus image fusion is to retain clear detail information of objects at different DoFs. We propose the noise transformation to generate locally blurred images for the encoder-decoder network to reconstruct, so that the trained model can learn to reconstruct clear images from locally blurred multi-focus source images.

We implement the noise transformation using Gaussian blur. Mathematically, applying Gaussian blur to an image is the same as convolving the image with a Gaussian function \cite{63}. In a two-dimensional image, the Gaussian function is defined as
\begin{equation}
    G\left(x,y\right)=\frac{1}{2\pi\sigma^2}e^{-\frac{x^2+y^2}{2\sigma^2}}
    \label{eq17}
\end{equation}
\begin{equation}
    {\widetilde{x}}_i\ =\ f_{{Ns}}\left(x_i\right)=x_i\ast G
    \label{eq18}
\end{equation}
where $x$ and $y$ are the distances of a point from the origin on the horizontal and vertical axis, respectively. $\sigma$ is the standard deviation of the Gaussian distribution, which we empirically set as three. Fig. \ref{figure2} (c) shows an original image subregion and the subregion after Gaussian blur transform.

\begin{figure*}[htbp]
    \centering
    \includegraphics[scale=0.6]{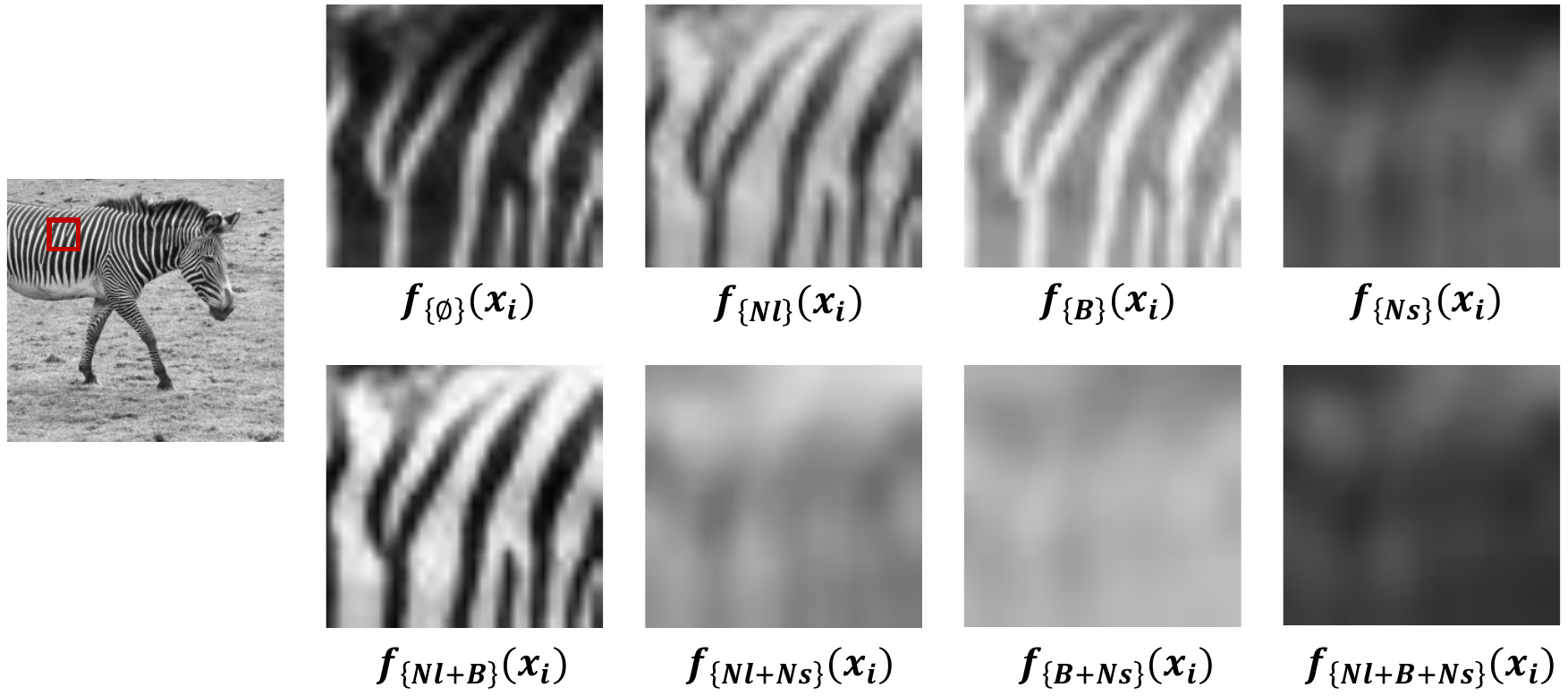}
    \caption{We integrate the three transformations designed for different fusion tasks using a probability-based combination strategy, enabling the network to learn task-specific features while extracting more generalizable features. Eight kinds of combinations of an image subregion are shown in the figure.}
    \label{figure3}
\end{figure*}

\subsubsection{Integrating the Three Transformations in a Unified Framework}
We have proposed three task-specific image transformations and our experiments showed that each of them can help improve the fusion performance of the corresponding image fusion task. Here we propose to integrate the three transformations in a probability-based combination strategy, so that the network can extract more generalized features and the fusion performance of each single task can be further improved. Concretely, for a subregion${\ x}_{i\ }$we apply the three transformations in the order of pixel intensity non-linear transformation, brightness transformation and noise transformation, but for each transformation we randomly decide whether to apply or skip it. Thus, we obtain eight possible combinations of transformations applied on an image patch, as illustrated in Fig. \ref{figure3}. In this way, the trained model can handle more diverse input images, including the original unchanged images, images transformed by one transformation or images transformed by two or even three different transformations.
The pseudo code for implementation is shown in Algorithm \ref{Algorithm1}.
\begin{algorithm}[ht]
    \SetAlgoLined
    \DontPrintSemicolon
    \KwIn{subregion to be transformed $x_i$, hyperparameter probability for each task-specific transformation ${Pro}_{Nl}$,\ ${Pro}_B$, ${Pro}_{Ns}$}    
    \KwOut{transformed subregion ${\widetilde{x}}_i$}
  
    Generate $r_{Nl}$ uniformly distributed over [0, 1) for non-linear transformation.

    \uIf {$r_{Nl}$ \textless ${Pro}_{Nl}$} {
        ${\widetilde{x}}_i \gets f_{\left\{Nl\right\}}(x_i)$
    } 
    \textbf{end}

    Generate $r_B$ uniformly distributed over [0, 1) for brightness transformation.

    \uIf {$r_B$ \textless ${Pro}_B$} {
        ${\widetilde{x}}_i \gets f_{\left\{B\right\}}({\widetilde{x}}_i)$
    } 
    \textbf{end}

    Generate $r_{Ns}$ uniformly distributed over [0, 1) for noise transformation.

    \uIf {$r_{Ns}$ \textless ${Pro}_{Ns}$} {
        ${\widetilde{x}}_i \gets f_{\left\{Ns\right\}}({\widetilde{x}}_i)$
    } 
    \textbf{end}
  
    \caption{Integrate three task-specific transformations by possible combinations in a Python-like style}
    \label{Algorithm1}
  \end{algorithm}

\subsection{Fusion Rule}
Due to the strong feature extraction capability of our network, fairly simple fusion rules can achieve very good fusion results. For multi-exposure image fusion task and multi-focus image fusion task, we directly average the feature maps of the two source images to obtain the fused feature maps. For multi-model image fusion, we adopt the L1-Norm fusion rule used in Li et al. \cite{33}, which can highlight and preserve the critical feature information in the fused feature maps adaptively according to the region energy in the feature maps.

\section{Experimental Results}
\subsection{Datasets}
We used the large natural image dataset MS-COCO \cite{64} to train the encoder-decoder network, which contains more than 70,000 natural images of various scenes. All images were resized to 256 $\times$ 256 and converted to grayscale images.

We used the following datasets to evaluate our image fusion framework and compare it to other methods in different types of image fusion tasks. In multi-modal image fusion, we used the TNO\footnote{https://figshare.com/articles/TNOImageFusionDataset/1008029} dataset for infrared and visible image fusion and the Harvard\footnote{http://www.med.harvard.edu/AANLIB/home.html} dataset for multi-modal medical image fusion. For multi-exposure image fusion, we used the dataset in \cite{65}, and for the multi-focus image fusion we used the dataset Lytro\footnote{https://mansournejati.ece.iut.ac.ir/content/lytro-multi-focus-dataset}. From each dataset, we randomly select 20 pairs of source images for testing. 

\subsection{Implementation Details}
Our model was trained on an NVIDIA GTX 3090 GPU with a batchsize of 64, epoch of 70, using an Adam optimizer and a cosine annealing learning rate adjustment strategy with a learning rate of 1e-4 and a weight decay of 5e-4. Given a 256 $\times$ 256 training image, we randomly generate four image subregions with the size of 16 $\times$ 16 to form the set $\chi$ to be transformed. In TransBlock, we divided the transformed input image into 256 patches with the size of 16×16 and constructed the sequence${\ x}_{seq_G}$. For each patch, we further divided it into 16 sub-patches with the size of 4 $\times$ 4 and constructed the sequence $x_{seq_L}$. We set all the probabilities ${Pro}_{NL}$, ${\ Pro}_B$\ and${\ Pro}_{Ns}$ as 0.6.

\subsection{Evaluation Metrics}
In the current image fusion research, evaluating an image fusion algorithm is not a simple task due to the lack of ground truth fusion results. There are two widely adopted methods for evaluating the fused images, namely subjective evaluation and objective evaluation \cite{66}. Subjective evaluation assesses the fused image in terms of sharpness, luminance, and contrast, etc. from the perspective of the observer. Objective evaluation assesses the fused images through objective evaluation metrics, but there is no consensus of the choice of evaluation metrics. Therefore, in order to provide a fair and comprehensive comparison with other fusion methods, we selected nine objective evaluation metrics focusing on four different aspects of the fused images and compared our method with state-of-the-art conventional and deep learning methods in each fusion task. 

The evaluation metrics includes: Information theory-based metrics: $\rm{FMI}$ \cite{58}, $\rm{Q^{NICE}}$ \cite{56}, $\rm{Q^{M}}$ \cite{81}; Image feature-based metrics: $\rm{Q^{A/BF}}$ \cite{59}, $\rm{Q^P}$ \cite{60}; Image structural similarity-based metrics: $\rm{SSIM}$ \cite{62}, $\rm{Q^Y}$ \cite{40}; Human perception inspired metrics: $\rm{VIF}$ \cite{63}, $\rm{Q^{CV}}$ \cite{80}. Among these quantitative evaluation metrics, the minimum value of $\rm{Q^{CV}}$ indicates the best fusion performance, while the maximum value indicates the best performance for all the other metrics. In every fusion task, the average of the objective metrics in fusion 20 image pairs were reported.

\subsection{Results}

\subsubsection{Comparation with Unified Image Fusion Framework}
Our model is a unified image fusion framework, so we first compared it with existing state-of-the-art unified image fusion algorithms U2Fusion \cite{17}, IFCNN \cite{18}, and PMGI \cite{19} in all the four image fusion tasks. The objective metrics of unified fusion methods are shown in Table \ref{table1}, where the four different tasks infrared-visible image fusion, multi-modal medical image fusion, multi-exposure image fusion and multi-focus image fusion are denoted as IV, MED, ME and MF, respectively. It is apparent from Table \ref{table1} that our method achieves the best fusion performance in almost all metrics. In subsequent experimental comparisons on each task, the three unified fusion methods are still included, so the comparison with their subjective fusion results is discussed in detail in the following sections.

\begin{table}[htbp]
    \centering
    \caption{Comparison of objective evaluations for unified image fusion tasks. Bolded red and bolded blue are used to denote the best and the second best values, respectively.}
    \scalebox{0.66}{
      \begin{tabular}{ccccccccccc}
      \toprule
      \rowcolor[rgb]{ .859,  .859,  .859} Task  & Method & $\rm{FMI}$ & $\rm{Q^{NICE}}$ & $\rm{Q^{M}}$ & $\rm{Q^{A/BF}}$ & $\rm{Q^P}$ & $\rm{SSIM}$ & $\rm{Q^Y}$ & $\rm{VIF}$ & $\rm{Q^{CV}}\downarrow$ \\
      \midrule
      \multirow{4}[4]{*}{\textit{IV}} & U2Fusion & 0.3546  & \textcolor[rgb]{ 0,  .439,  .753}{\textbf{0.8055}} & 0.4380  & 0.4502  & \textcolor[rgb]{ 0,  .439,  .753}{\textbf{0.3331}} & \textcolor[rgb]{ 1,  0,  0}{\textbf{0.9196}} & 0.7161  & \textcolor[rgb]{ 1,  0,  0}{\textbf{0.5637}} & 451.9610  \\
            & IFCNN & \textcolor[rgb]{ 0,  .439,  .753}{\textbf{0.3985}} & 0.8050  & \textcolor[rgb]{ 0,  .439,  .753}{\textbf{0.5193}} & \textcolor[rgb]{ 0,  .439,  .753}{\textbf{0.4670}} & 0.3212  & 0.8726  & \textcolor[rgb]{ 0,  .439,  .753}{\textbf{0.7574}} & 0.2701  & \textcolor[rgb]{ 0,  .439,  .753}{\textbf{436.1472}} \\
            & PMGI  & 0.3510  & 0.8049  & 0.4022  & 0.3461  & 0.1862  & 0.7471  & 0.6164  & \textcolor[rgb]{ 0,  .439,  .753}{\textbf{0.4896}} & 706.8626  \\
  \cmidrule{2-11}          & \textbf{ours} & \textcolor[rgb]{ 1,  0,  0}{\textbf{0.4223}} & \textcolor[rgb]{ 1,  0,  0}{\textbf{0.8083}} & \textcolor[rgb]{ 1,  0,  0}{\textbf{0.5773}} & \textcolor[rgb]{ 1,  0,  0}{\textbf{0.4986}} & \textcolor[rgb]{ 1,  0,  0}{\textbf{0.3689}} & \textcolor[rgb]{ 0,  .439,  .753}{\textbf{0.8733}} & \textcolor[rgb]{ 1,  0,  0}{\textbf{0.7605}} & 0.4054  & \textcolor[rgb]{ 1,  0,  0}{\textbf{280.1503}} \\
      \midrule
      \multirow{4}[4]{*}{\textit{MED}} & U2Fusion & 0.3471  & 0.8067  & 0.1664  & 0.4658  & 0.3550  & 0.8980  & 0.4174  & 0.4535  & 1166.1936  \\
            & IFCNN & \textcolor[rgb]{ 1,  0,  0}{\textbf{0.4192}} & \textcolor[rgb]{ 0,  .439,  .753}{\textbf{0.8071}} & \textcolor[rgb]{ 0,  .439,  .753}{\textbf{0.2094}} & \textcolor[rgb]{ 0,  .439,  .753}{\textbf{0.5232}} & \textcolor[rgb]{ 0,  .439,  .753}{\textbf{0.3797}} & \textcolor[rgb]{ 1,  0,  0}{\textbf{0.9183}} & \textcolor[rgb]{ 0,  .439,  .753}{\textbf{0.4634}} & 0.5134  & \textcolor[rgb]{ 1,  0,  0}{\textbf{933.5277}} \\
            & PMGI  & 0.2993  & 0.8068  & 0.1423  & 0.2756  & 0.2326  & 0.8188  & 0.3469  & \textcolor[rgb]{ 0,  .439,  .753}{\textbf{0.5166}} & 1532.5396  \\
  \cmidrule{2-11}          & \textbf{ours} & \textcolor[rgb]{ 0,  .439,  .753}{\textbf{0.3604}} & \textcolor[rgb]{ 1,  0,  0}{\textbf{0.8094}} & \textcolor[rgb]{ 1,  0,  0}{\textbf{0.2301}} & \textcolor[rgb]{ 1,  0,  0}{\textbf{0.5275}} & \textcolor[rgb]{ 1,  0,  0}{\textbf{0.4620}} & \textcolor[rgb]{ 0,  .439,  .753}{\textbf{0.9046}} & \textcolor[rgb]{ 1,  0,  0}{\textbf{0.5073}} & \textcolor[rgb]{ 1,  0,  0}{\textbf{0.5201}} & \textcolor[rgb]{ 0,  .439,  .753}{\textbf{936.7991}} \\
      \midrule
      \multirow{4}[4]{*}{\textit{ME}} & U2Fusion & 0.4411  & 0.8146  & 0.4436  & 0.6146  & \textcolor[rgb]{ 0,  .439,  .753}{\textbf{0.7055}} & 0.9349  & 0.6340  & \textcolor[rgb]{ 1,  0,  0}{\textbf{1.2437}} & \textcolor[rgb]{ 1,  0,  0}{\textbf{174.8057}} \\
            & IFCNN & \textcolor[rgb]{ 0,  .439,  .753}{\textbf{0.5058}} & 0.8141  & \textcolor[rgb]{ 1,  0,  0}{\textbf{0.6753}} & \textcolor[rgb]{ 0,  .439,  .753}{\textbf{0.6890}} & 0.6926  & \textcolor[rgb]{ 0,  .439,  .753}{\textbf{0.9445}} & \textcolor[rgb]{ 1,  0,  0}{\textbf{0.7366}} & 0.5889  & 192.5404  \\
            & PMGI  & 0.4157  & 0.8150  & 0.4119  & 0.4578  & 0.5088  & 0.9083  & 0.5981  & \textcolor[rgb]{ 0,  .439,  .753}{\textbf{0.7557}} & 297.7641  \\
  \cmidrule{2-11}          & \textbf{ours} & \textcolor[rgb]{ 1,  0,  0}{\textbf{0.5340}} & \textcolor[rgb]{ 1,  0,  0}{\textbf{0.8271}} & \textcolor[rgb]{ 0,  .439,  .753}{\textbf{0.6703}} & \textcolor[rgb]{ 1,  0,  0}{\textbf{0.7322}} & \textcolor[rgb]{ 1,  0,  0}{\textbf{0.7665}} & \textcolor[rgb]{ 1,  0,  0}{\textbf{0.9579}} & \textcolor[rgb]{ 0,  .439,  .753}{\textbf{0.7118}} & \textcolor[rgb]{ 0,  .439,  .753}{\textbf{1.1042}} & \textcolor[rgb]{ 0,  .439,  .753}{\textbf{192.2550}} \\
      \midrule
      \multirow{4}[4]{*}{\textit{MF}} & U2Fusion & 0.4653  & 0.8269  & 0.4650  & 0.6637  & 0.7936  & 0.9574  & 0.8487  & \textcolor[rgb]{ 1,  0,  0}{\textbf{1.1727}} & 102.8688  \\
            & IFCNN & \textcolor[rgb]{ 0,  .439,  .753}{\textbf{0.5438}} & \textcolor[rgb]{ 0,  .439,  .753}{\textbf{0.8287}} & \textcolor[rgb]{ 0,  .439,  .753}{\textbf{0.5845}} & \textcolor[rgb]{ 0,  .439,  .753}{\textbf{0.7103}} & \textcolor[rgb]{ 0,  .439,  .753}{\textbf{0.8317}} & \textcolor[rgb]{ 0,  .439,  .753}{\textbf{0.9848}} & \textcolor[rgb]{ 0,  .439,  .753}{\textbf{0.8971}} & 0.9915  & \textcolor[rgb]{ 0,  .439,  .753}{\textbf{64.0155}} \\
            & PMGI  & 0.4366  & 0.8255  & 0.3860  & 0.4610  & 0.5417  & 0.8532  & 0.6555  & \textcolor[rgb]{ 0,  .439,  .753}{\textbf{0.9405}} & 373.6547  \\
  \cmidrule{2-11}          & \textbf{ours} & \textcolor[rgb]{ 1,  0,  0}{\textbf{0.5620}} & \textcolor[rgb]{ 1,  0,  0}{\textbf{0.8367}} & \textcolor[rgb]{ 1,  0,  0}{\textbf{0.6666}} & \textcolor[rgb]{ 1,  0,  0}{\textbf{0.7560}} & \textcolor[rgb]{ 1,  0,  0}{\textbf{0.8374}} & \textcolor[rgb]{ 1,  0,  0}{\textbf{0.9878}} & \textcolor[rgb]{ 1,  0,  0}{\textbf{0.9049}} & \textcolor[rgb]{ 0,  .439,  .753}{\textbf{0.9935}} & \textcolor[rgb]{ 1,  0,  0}{\textbf{56.4455}} \\
      \bottomrule
      \end{tabular}%
    }
    \label{table1}%
  \end{table}%
  
\subsubsection{Multi-Modal Image Fusion}
\begin{figure*}[htbp]
    \centering
    \includegraphics[scale=0.49]{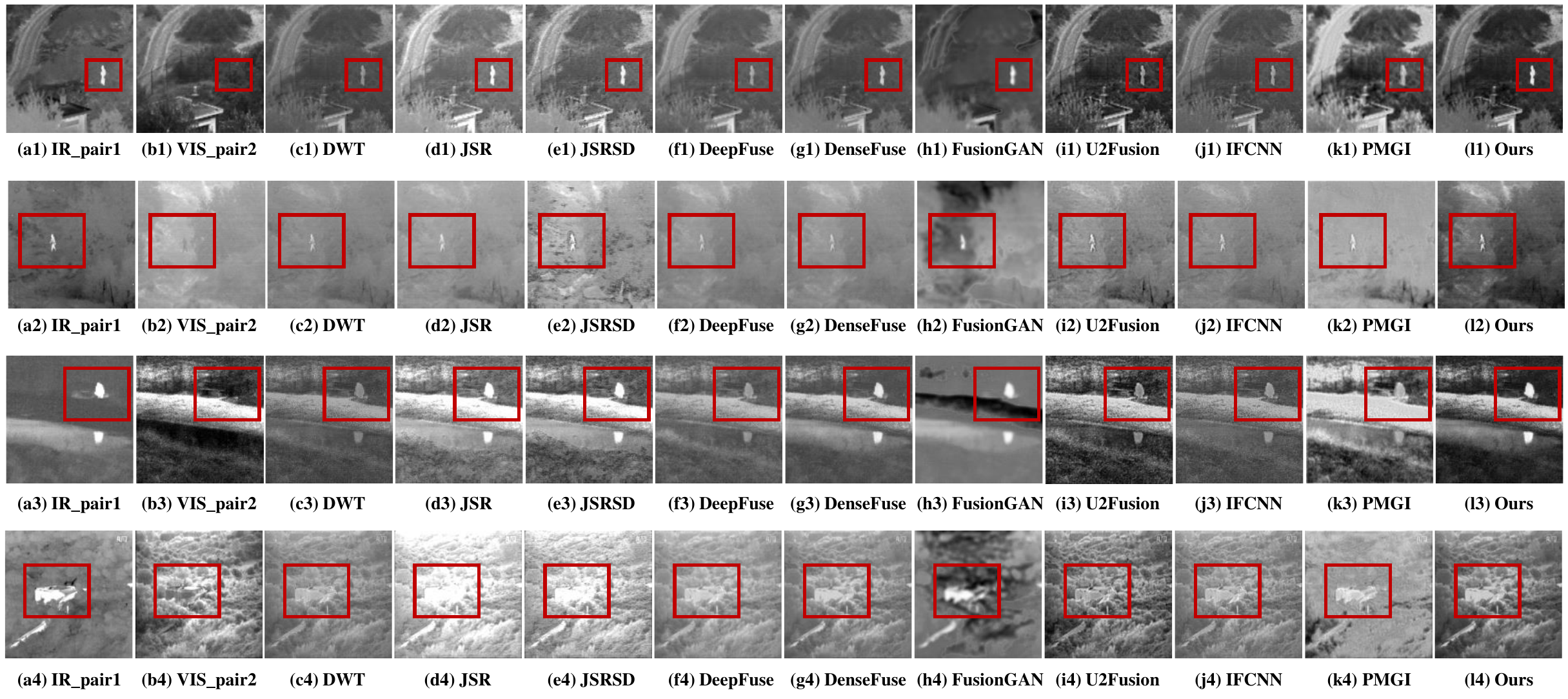}
    \caption{Infrared and visible source images and their fusion results, where (a1)-(b1), (a2)-(b2), (a3)-(b3), (a4)-(b4) are the infrared visible source image pairs respectively. (c1)-(l1), (c2)-(l2), (c3)-(l3), (c4)-(l4) are the fusion results of the comparative methods.}
    \label{figure4}
\end{figure*}

\paragraph{Visible and Infrared Image Fusion}
We compared our method with nine representative methods in infrared and visible image fusion task, including traditional methods (DWT \cite{1}, JSR \cite{29}, JSRSD \cite{29}) and deep learning-based methods (U2Fusion \cite{17}, DeepFuse \cite{32}, DenseFuse \cite{33}, FusionGAN \cite{34}, IFCNN \cite{18}, PMGI \cite{19}). The fusion results are shown in Fig. \ref{figure4}, and the comparisons of objective evaluation metrics are shown in Table \ref{table2}.

\subparagraph{Subjective Evaluation}
In Fig. \ref{figure4}, JSR, JSRSD, FusionGAN and PMGI show relatively disappointing fusion results due to noise and artifacts. DWT, DeepFuse and DenseFuse show relatively similar fusion results, where the fused images have low contrast and blurred details and blurred edges. U2Fusion, IFCNN and our method all achieve satisfying fusion results, while our method preserves sharper texture details with higher contrast and sharpness. As can be seen from the red box in Fig. \ref{figure4}, our method highlights the key targets (the humans in (a1)-(b1), (a2)-(b2) and (a3)-(b3) and the house in (a4)-(b4)) and has the best visual effect.

\subparagraph{Objective Evaluation}
As can be seen in Table \ref{table2}, our method achieves the best results in four metrics and the second best results in other four metrics among all nine objective metrics, which is the best in all comparing methods. DenseFuse is specially designed for this task and it also shows strong performance.

\begin{table}[htbp]
    \centering
    \caption{Comparison of objective metrics on the infrared and visible image fusion dataset. The best values are bolded in red the second best values are bolded in blue.}
    \scalebox{0.66}{
      \begin{tabular}{ccccccccccc}
      \toprule
      \rowcolor[rgb]{ .859,  .859,  .859} Task  & Method & $\rm{FMI}$ & $\rm{Q^{NICE}}$ & $\rm{Q^{M}}$ & $\rm{Q^{A/BF}}$ & $\rm{Q^P}$ & $\rm{SSIM}$ & $\rm{Q^Y}$ & $\rm{VIF}$ & $\rm{Q^{CV}}\downarrow$ \\
      \midrule
      \multirow{10}[6]{*}{\textit{IV}} & DWT   & 0.3544  & 0.8050  & 0.4991  & 0.4005  & 0.3038  & 0.8401  & 0.7051  & 0.2321  & 457.8833  \\
            & JSR   & 0.2166  & 0.8055  & 0.3754  & 0.4060  & 0.2121  & 0.8475  & 0.5992  & 0.3448  & 350.9396  \\
            & JSRSD & 0.1924  & 0.8056  & 0.3089  & 0.3746  & 0.1524  & 0.7636  & 0.5219  & 0.2778  & 391.0290  \\
  \cmidrule{2-11}          & U2Fusion & 0.3546  & 0.8055  & 0.4380  & 0.4502  & 0.3331  & \textcolor[rgb]{ 1,  0,  0}{\textbf{0.9196}} & 0.7161  & \textcolor[rgb]{ 1,  0,  0}{\textbf{0.5637}} & 451.9610  \\
            & DeepFuse & 0.3718  & 0.8051  & 0.4317  & 0.3746  & 0.3034  & 0.8473  & 0.6782  & 0.2523  & 458.7481  \\
            & DenseFuse & \textcolor[rgb]{ 0,  .439,  .753}{\textbf{0.4157}} & \textcolor[rgb]{ 1,  0,  0}{\textbf{0.8083}} & \textcolor[rgb]{ 0,  .439,  .753}{\textbf{0.5237}} & \textcolor[rgb]{ 0,  .439,  .753}{\textbf{0.4754}} & \textcolor[rgb]{ 1,  0,  0}{\textbf{0.3740}} & 0.8131  & \textcolor[rgb]{ 1,  0,  0}{\textbf{0.7885}} & 0.1978  & \textcolor[rgb]{ 1,  0,  0}{\textbf{275.0618}} \\
            & FusionGAN & 0.2937  & \textcolor[rgb]{ 0,  .439,  .753}{\textbf{0.8080}} & 0.3355  & 0.1923  & 0.0859  & 0.3963  & 0.3362  & 0.0797  & 2291.7420  \\
            & IFCNN & 0.3985  & 0.8050  & 0.5193  & 0.4670  & 0.3212  & 0.8726  & 0.7574  & 0.2701  & 436.1472  \\
            & PMGI  & 0.3510  & 0.8049  & 0.4022  & 0.3461  & 0.1862  & 0.7471  & 0.6164  & \textcolor[rgb]{ 0,  .439,  .753}{\textbf{0.4896}} & 706.8626  \\
  \cmidrule{2-11}          & \textbf{ours} & \textcolor[rgb]{ 1,  0,  0}{\textbf{0.4223}} & \textcolor[rgb]{ 1,  0,  0}{\textbf{0.8083}} & \textcolor[rgb]{ 1,  0,  0}{\textbf{0.5773}} & \textcolor[rgb]{ 1,  0,  0}{\textbf{0.4986}} & \textcolor[rgb]{ 0,  .439,  .753}{\textbf{0.3689}} & \textcolor[rgb]{ 0,  .439,  .753}{\textbf{0.8733}} & \textcolor[rgb]{ 0,  .439,  .753}{\textbf{0.7605}} & 0.4054  & \textcolor[rgb]{ 0,  .439,  .753}{\textbf{280.1503}} \\
      \bottomrule
      \end{tabular}%
    \label{table2}%
    }
  \end{table}%

\begin{figure*}[htbp]
    \centering
    \includegraphics[scale=0.65]{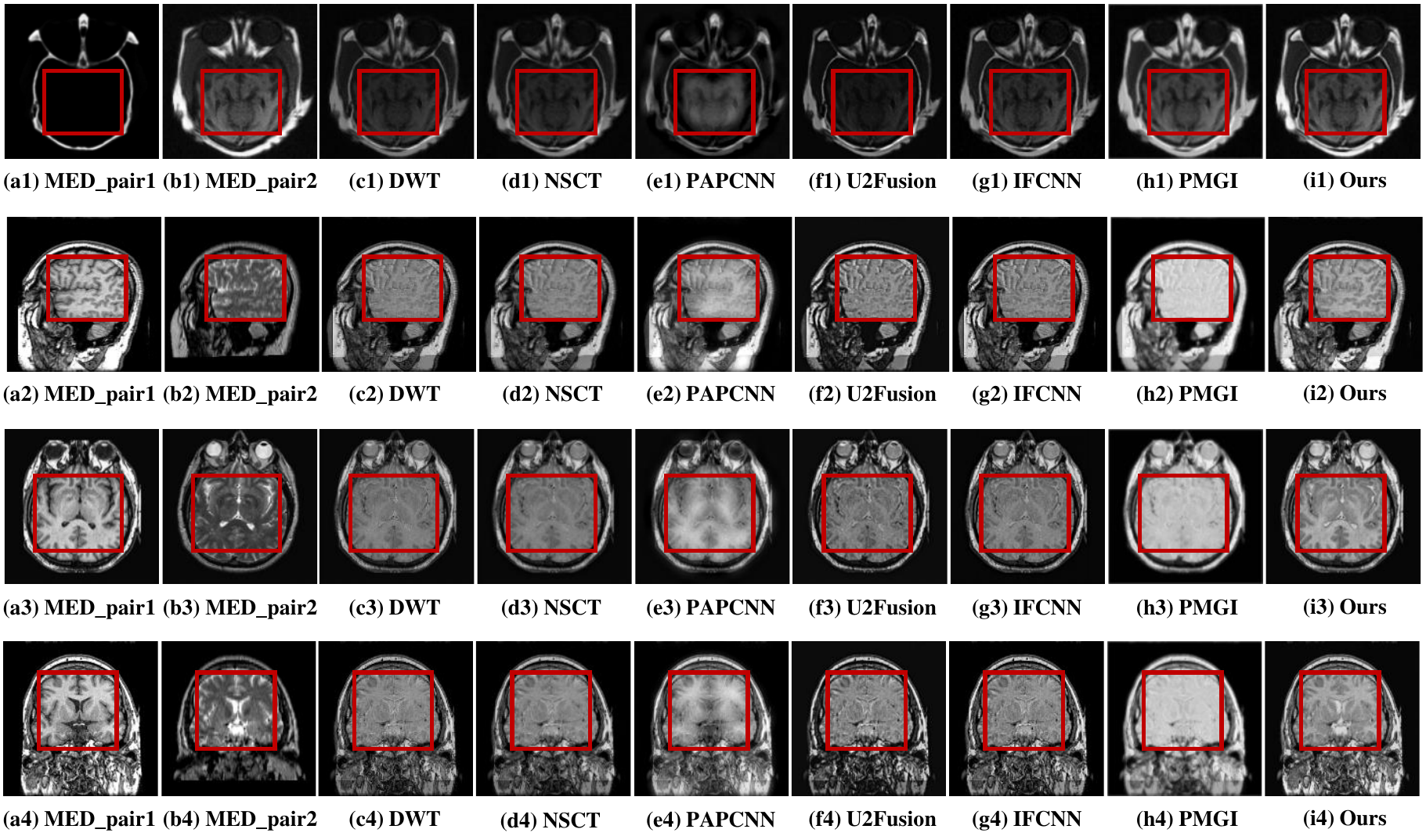}
    \caption{Medical source images and their fusion results, where (a1)-(b1) are the medical source image pairs in "CT" and "MRI" modalities, (a2)-(b2), (a3)-(b3), (a4)-(b4) are the medical source image pairs in "T1" and "T2" modalities. (c1)-(i1), (c2)-(i2), (c3)-(i3) and (c4)-(i4) are the results of the comparative methods.}
    \label{figure5}
\end{figure*}

\paragraph{Medical Image Fusion}
We compared our method with five current state-of-the-art methods in medical image fusion task, which contain the traditional methods DWT \cite{1}, NSCT \cite{73}, PAPCNN \cite{74} and the deep learning methods U2Fusion \cite{17}, IFCNN \cite{18}, PMGI \cite{19}. The fusion results are shown in Fig. \ref{figure5}, and the comparison of objective evaluation metrics are shown in Table \ref{table3}.

\subparagraph{Subjective Evaluation}
The fusion results in Fig. \ref{figure5} show that DWT, NSCT and PMGI obtain blurred fused images with low contrast and sharpness. PAPCNN introduces severe artifacts and noise. Although U2Fusion and IFCNN enhance the edges and detail information, their fused images have low contrast and inferior visual effect. In contrast, our method retains the best texture and functional information, while showing the optimal contrast and details.
  
\subparagraph{Objective Evaluation}
As can be seen in Table \ref{table3}, our method achieves the best performance in six metrics and the second best performance in three metrics with a small difference from the best ones on the medical Harvard test dataset. In general, our method achieves the best fusion performance.

\begin{table}[htbp]
    \centering
    \caption{Comparison of objective metrics on the medical image fusion dataset. The best values are bolded in red the second best values are bolded in blue.}
    \scalebox{0.66}{
      \begin{tabular}{ccccccccccc}
      \toprule
      \rowcolor[rgb]{ .859,  .859,  .859} Task  & Method & $\rm{FMI}$ & $\rm{Q^{NICE}}$ & $\rm{Q^{M}}$ & $\rm{Q^{A/BF}}$ & $\rm{Q^P}$ & $\rm{SSIM}$ & $\rm{Q^Y}$ & $\rm{VIF}$ & $\rm{Q^{CV}}\downarrow$ \\
      \midrule
      \multirow{6}[6]{*}{\textit{MED}} & DWT   & 0.3588  & \textcolor[rgb]{ 0,  .439,  .753}{\textbf{0.8072}} & 0.1858  & 0.4336  & 0.3409  & 0.8827  & \textcolor[rgb]{ 0,  .439,  .753}{\textbf{0.4960}} & 0.4684  & 1143.3950  \\
            & PAPCNN & 0.3518  & 0.8066  & 0.1634  & 0.4821  & 0.3299  & 0.8796  & 0.4250  & 0.5039  & 1078.2390  \\
  \cmidrule{2-11}          & U2Fusion & 0.3471  & 0.8067  & 0.1664  & 0.4658  & 0.3550  & 0.8980  & 0.4174  & 0.4535  & 1166.1936  \\
            & IFCNN & \textcolor[rgb]{ 1,  0,  0}{\textbf{0.4192}} & 0.8071  & \textcolor[rgb]{ 0,  .439,  .753}{\textbf{0.2094}} & \textcolor[rgb]{ 0,  .439,  .753}{\textbf{0.5232}} & \textcolor[rgb]{ 0,  .439,  .753}{\textbf{0.3797}} & \textcolor[rgb]{ 1,  0,  0}{\textbf{0.9183}} & 0.4634  & 0.5134  & \textcolor[rgb]{ 1,  0,  0}{\textbf{933.5277}} \\
            & PMGI  & 0.2993  & 0.8068  & 0.1423  & 0.2756  & 0.2326  & 0.8188  & 0.3469  & \textcolor[rgb]{ 0,  .439,  .753}{\textbf{0.5166}} & 1532.5396  \\
  \cmidrule{2-11}          & \textbf{ours} & \textcolor[rgb]{ 0,  .439,  .753}{\textbf{0.3604}} & \textcolor[rgb]{ 1,  0,  0}{\textbf{0.8094}} & \textcolor[rgb]{ 1,  0,  0}{\textbf{0.2301}} & \textcolor[rgb]{ 1,  0,  0}{\textbf{0.5275}} & \textcolor[rgb]{ 1,  0,  0}{\textbf{0.4620}} & \textcolor[rgb]{ 0,  .439,  .753}{\textbf{0.9046}} & \textcolor[rgb]{ 1,  0,  0}{\textbf{0.5073}} & \textcolor[rgb]{ 1,  0,  0}{\textbf{0.5201}} & \textcolor[rgb]{ 0,  .439,  .753}{\textbf{936.7991}} \\
      \bottomrule
      \end{tabular}%
    \label{table3}%
    }
  \end{table}%
  
\begin{figure*}[htbp]
    \centering
    \includegraphics[scale=0.65]{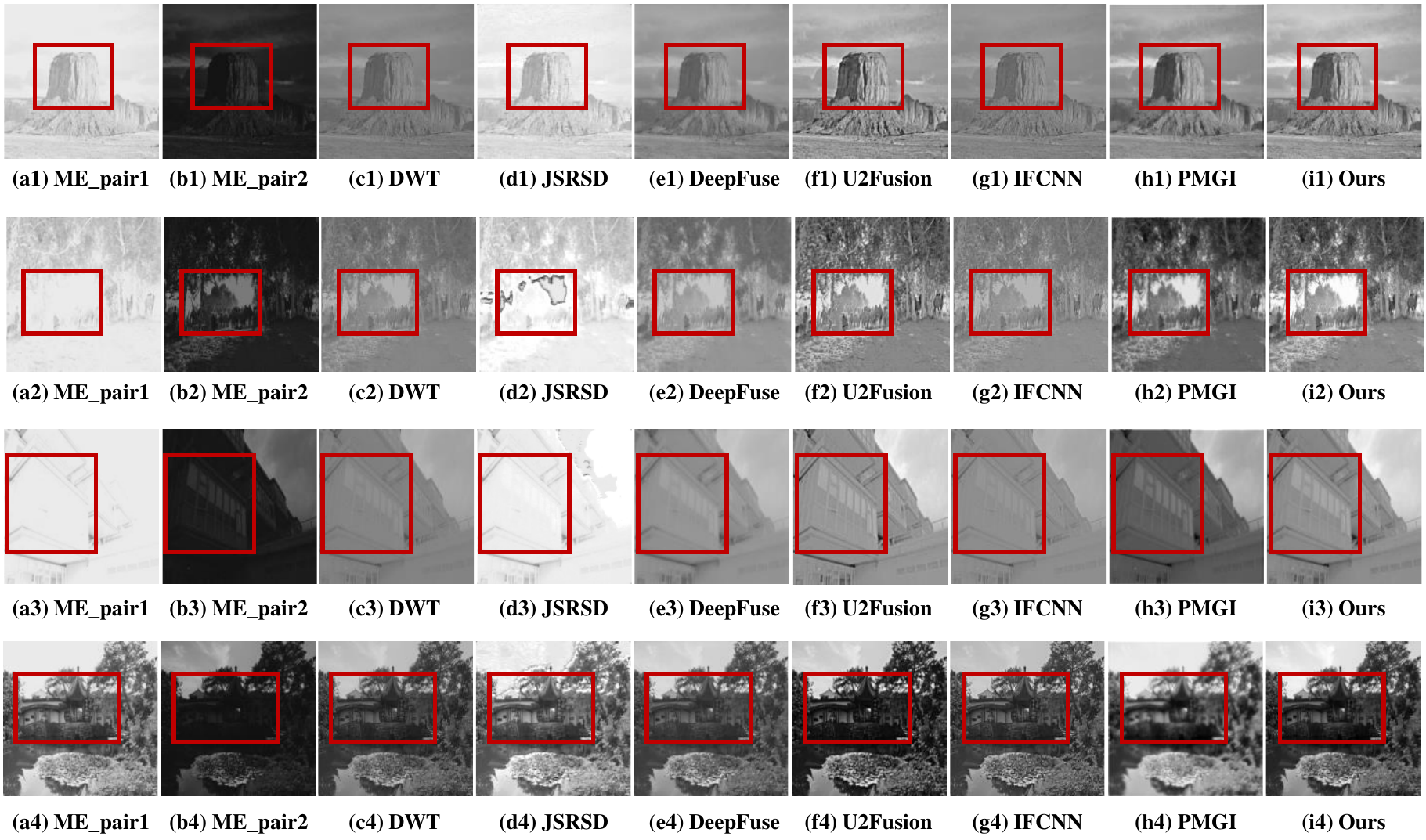}
    \caption{Multi-exposure source images and their fusion results, where (a1)-(b1), (a2)-(b2), (a3)-(b3), (a4)-(b4) are multi-exposure source image pairs and (c1)-(i1), (c2)-(i2), (c3)-(i3), (c4)-(i4) are fusion results of the comparative methods respectively.}
    \label{figure6}
\end{figure*}

\paragraph{Multi-Exposure Image Fusion}
We compared our method with five current state-of-the-art methods in multi-exposure image fusion task, including traditional methods (DWT \cite{1}, JSRSD \cite{29}) and deep learning-based methods (DeepFuse \cite{32}, U2Fusion \cite{17}, IFCNN \cite{18}, PMGI \cite{19}). The fusion results are shown in Fig. \ref{figure6}, and the comparisons of objective evaluation metrics are shown in Table \ref{table4}.

\subparagraph{Subjective Evaluation}
Fig. \ref{figure6} show that JSRSD introduces severe noise and artifacts. DWT, DeepFuse and IFCNN fail to maintain appropriate luminance with contrast and details lost. U2Fusion, PMGI and our method achieve better fusion results. However, U2Fusion over-sharpens the details and edge information, resulting in unnatural visual effect and PMGI lose more details. In contrast, our fused images maintain the best luminance and have high contrast and sharpness. Overall, our fused images are more natural and achieved optimal visual effect.

\subparagraph{Objective Evaluation}
As can be seen in Table \ref{table4}, our method achieves the best performance in five metrics and the second best performance in other four metrics on the multi-exposure image fusion dataset. In comparison, our method achieves the best fusion performance.

\begin{table}[htbp]
    \centering
    \caption{Comparison of objective metrics on the multi-exposure image fusion dataset. The best values are bolded in red the second best values are bolded in blue.}
    \scalebox{0.66}{
      \begin{tabular}{ccccccccccc}
      \toprule
      \rowcolor[rgb]{ .859,  .859,  .859} Task  & Method & $\rm{FMI}$ & $\rm{Q^{NICE}}$ & $\rm{Q^{M}}$ & $\rm{Q^{A/BF}}$ & $\rm{Q^P}$ & $\rm{SSIM}$ & $\rm{Q^Y}$ & $\rm{VIF}$ & $\rm{Q^{CV}}\downarrow$ \\
      \midrule
      \multirow{7}[6]{*}{\textit{ME}} & DWT   & 0.4727  & \textcolor[rgb]{ 0,  .439,  .753}{\textbf{0.8174}} & 0.6077  & 0.5978  & 0.6601  & 0.9227  & 0.6933  & 0.4952  & 223.4195  \\
            & JSRSD & 0.3026  & 0.8150  & 0.5428  & 0.6174  & 0.4461  & 0.9001  & 0.6771  & 0.8074  & 307.0123  \\
  \cmidrule{2-11}          & U2Fusion & 0.4411  & 0.8146  & 0.4436  & 0.6146  & \textcolor[rgb]{ 0,  .439,  .753}{\textbf{0.7055}} & 0.9349  & 0.6340  & \textcolor[rgb]{ 1,  0,  0}{\textbf{1.2437}} & \textcolor[rgb]{ 1,  0,  0}{\textbf{174.8057}} \\
            & DeepFuse & 0.4462  & 0.8163  & 0.4289  & 0.5253  & 0.6330  & 0.9115  & 0.6501  & 0.6306  & 217.6652  \\
            & IFCNN & \textcolor[rgb]{ 0,  .439,  .753}{\textbf{0.5058}} & 0.8141  & \textcolor[rgb]{ 1,  0,  0}{\textbf{0.6753}} & \textcolor[rgb]{ 0,  .439,  .753}{\textbf{0.6890}} & 0.6926  & \textcolor[rgb]{ 0,  .439,  .753}{\textbf{0.9445}} & \textcolor[rgb]{ 1,  0,  0}{\textbf{0.7366}} & 0.5889  & 192.5404  \\
            & PMGI  & 0.4157  & 0.8150  & 0.4119  & 0.4578  & 0.5088  & 0.9083  & 0.5981  & \textcolor[rgb]{ 0,  .439,  .753}{\textbf{0.7557}} & 297.7641  \\
  \cmidrule{2-11}          & \textbf{ours} & \textcolor[rgb]{ 1,  0,  0}{\textbf{0.5340}} & \textcolor[rgb]{ 1,  0,  0}{\textbf{0.8271}} & \textcolor[rgb]{ 0,  .439,  .753}{\textbf{0.6703}} & \textcolor[rgb]{ 1,  0,  0}{\textbf{0.7322}} & \textcolor[rgb]{ 1,  0,  0}{\textbf{0.7665}} & \textcolor[rgb]{ 1,  0,  0}{\textbf{0.9579}} & \textcolor[rgb]{ 0,  .439,  .753}{\textbf{0.7118}} & \textcolor[rgb]{ 0,  .439,  .753}{\textbf{1.1042}} & \textcolor[rgb]{ 0,  .439,  .753}{\textbf{192.2550}} \\
      \bottomrule
      \end{tabular}%
    \label{table4}%
    }
  \end{table}%
  
\begin{figure*}[htbp]
    \centering
    \includegraphics[scale=0.59]{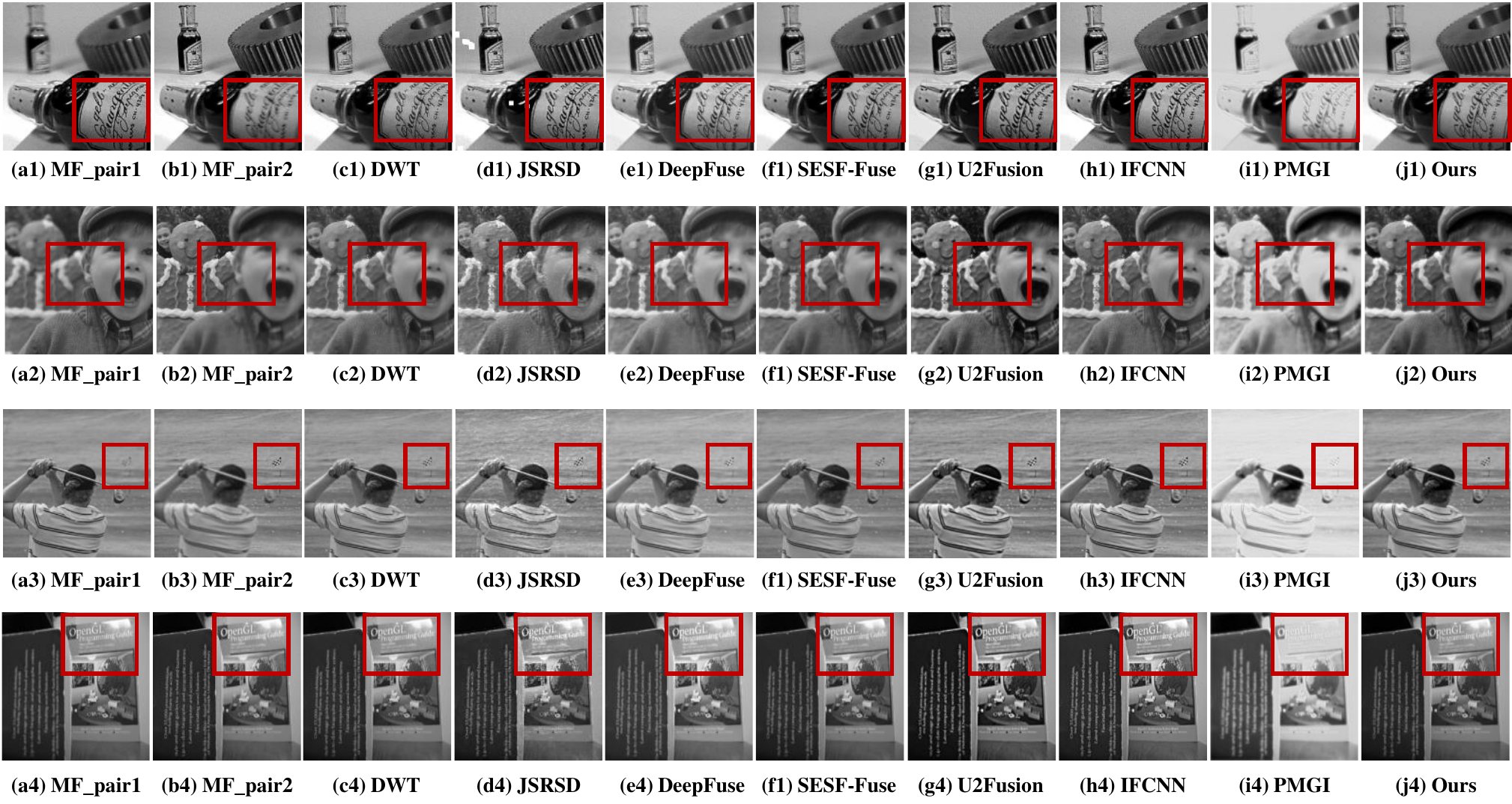}
    \caption{Multi-focus source images and their fusion results, where (a1)-(b1), (a2)-(b2), (a3)-(b3), (a4)-(b4) are multi-focus source image pairs and (c1)-(j1), (c2)-(j2), (c3)-(j3), (c4)-(j4) are fusion results of the comparative methods.}
    \label{figure7}
\end{figure*}

  \paragraph{Multi-Focus Image Fusion}
  We compared our method with six representative methods in multi-focus image fusion task, including traditional methods (DWT \cite{1}, JSRSD \cite{29}) and deep learning-based methods (DeepFuse \cite{32}, SESF-Fuse \cite{36}, U2Fusion \cite{17}, IFCNN \cite{18}, PMGI \cite{19}). The fusion results are shown in Fig. \ref{figure7}, and the comparison of objective evaluation metrics are shown in Table \ref{table5}.

\subparagraph{Subjective Evaluation}
From Fig. \ref{figure7}, JSRSD and PMGI introduces some noise and obtain unsatisfactory fusion results. Although DWT, DeepFuse and SESF-Fuse successfully fuse the images, they still lose detailed information, such as the flag or the edge contours of the child marked in red boxes, etc. U2Fusion, IFCNN and our method all achieves satisfying fusion results.

\subparagraph{Objective Evaluation}
We did not compare the objective metric of JSRSD owing to the artifacts in its fusion images. As can be seen in Table \ref{table5}, our method achieves the best performance in six metrics and the second best performance in other three metrics on the multi-exposure image fusion dataset. In comparison, our method achieves the best fusion performance.

\begin{table}[htbp]
    \centering
    \caption{Comparison of objective metrics on the multi-focus image fusion dataset. The best values are bolded in red the second best values are bolded in blue.}
    \scalebox{0.66}{
      \begin{tabular}{ccccccccccc}
      \toprule
      \rowcolor[rgb]{ .859,  .859,  .859} Task  & Method & $\rm{FMI}$ & $\rm{Q^{NICE}}$ & $\rm{Q^{M}}$ & $\rm{Q^{A/BF}}$ & $\rm{Q^P}$ & $\rm{SSIM}$ & $\rm{Q^Y}$ & $\rm{VIF}$ & $\rm{Q^{CV}}\downarrow$ \\
      \midrule
      \multirow{6}[4]{*}{\textit{MF}} & U2Fusion & 0.4653  & 0.8269  & 0.4650  & 0.6637  & 0.7936  & 0.9574  & 0.8487  & \textcolor[rgb]{ 1,  0,  0}{\textbf{1.1727 }} & 102.8688  \\
            & DeepFuse & 0.4868  & 0.8292  & 0.4960  & 0.7104  & 0.7757  & 0.9812  & 0.8608  & 0.9033  & 71.3387  \\
            & IFCNN & 0.5438  & 0.8287  & 0.5845  & 0.7103  & 0.8317  & 0.9848  & 0.8971  & 0.9915  & 64.0155  \\
            & SESF-Fuse & \textcolor[rgb]{ 0,  .439,  .753}{\textbf{0.5565}} & \textcolor[rgb]{ 0,  .439,  .753}{\textbf{0.8328}} & \textcolor[rgb]{ 1,  0,  0}{\textbf{0.7138}} & \textcolor[rgb]{ 0,  .439,  .753}{\textbf{0.7473}} & \textcolor[rgb]{ 0,  .439,  .753}{\textbf{0.8357}} & \textcolor[rgb]{ 0,  .439,  .753}{\textbf{0.9877}} & \textcolor[rgb]{ 0,  .439,  .753}{\textbf{0.9039}} & 0.8613  & \textcolor[rgb]{ 1,  0,  0}{\textbf{49.1273}} \\
            & PMGI  & 0.4366  & 0.8255  & 0.3860  & 0.4610  & 0.5417  & 0.8532  & 0.6555  & \textcolor[rgb]{ 0,  .439,  .753}{\textbf{0.9405}} & 373.6547  \\
  \cmidrule{2-11}          & \textbf{ours} & \textcolor[rgb]{ 1,  0,  0}{\textbf{0.5620}} & \textcolor[rgb]{ 1,  0,  0}{\textbf{0.8367}} & \textcolor[rgb]{ 0,  .439,  .753}{\textbf{0.6666}} & \textcolor[rgb]{ 1,  0,  0}{\textbf{0.7560}} & \textcolor[rgb]{ 1,  0,  0}{\textbf{0.8374}} & \textcolor[rgb]{ 1,  0,  0}{\textbf{0.9878}} & \textcolor[rgb]{ 1,  0,  0}{\textbf{0.9049}} & \textcolor[rgb]{ 0,  .439,  .753}{\textbf{0.9935}} & \textcolor[rgb]{ 0,  .439,  .753}{\textbf{56.4455}} \\
      \bottomrule
      \end{tabular}%
    \label{table5}%
    }
  \end{table}%
  
\section{Ablation Study}
\subsection{Ablation Study for TransBlock}
To solve the difficulties in establishing long-range relational dependencies of CNN-based image fusion architectures, we design TransBlock that combines CNN with Transformer to enable the network to exploit both local and global information. To verify the effectiveness of TransBlock, we perform ablation experiments on all fusion tasks using 20\% of the training data and the results are shown in Table \ref{table6}. “3 Transformations” in the table denotes that using proposed three task-specific image transformations and the probability-based combination strategy for training. As shown in Fig. \ref{figure1} (a), we remove the Transformer-Module in TransBlock along with the lower branch of the EnhanceBlock and use only the CNN-Module and the upper branch of the EnhanceBlock for comparison. As shown in Table \ref{table6}, almost in all fusion tasks, no matter whether the three transformations are used or not, adding TransBlock always improves the performance, which shows its effectiveness.

\begin{table}[htbp]
    \centering
    \caption{Results of the TransBlock ablation tests using 20\% training data}
    \scalebox{0.55}{
      \begin{tabular}{cccccccccccc}
      \toprule
      \rowcolor[rgb]{ .859,  .859,  .859} Task  & TransBlock & 3 Transformations & $\rm{FMI}$  & $\rm{Q^{NICE}}$ & $\rm{Q^{M}}$ & $\rm{Q^{A/BF}}$ & $\rm{Q^P}$ & $\rm{SSIM}$ & $\rm{Q^Y}$ & $\rm{VIF}$ & $\rm{Q^{CV}}\downarrow$ \\
      \midrule
      \multirow{4}[2]{*}{\textit{IV}} &       &       & 0.2775  & 0.8075  & 0.4669  & 0.4692  & 0.3265  & 0.8745  & 0.7158  & 0.3873  & 295.9092  \\
            & \checkmark     &       & 0.3135  & 0.8085  & 0.5172  & 0.4993  & 0.3606  & 0.8298  & 0.7773  & 0.3088  & 301.0812  \\
            &       & \checkmark     & 0.3204  & 0.8079  & 0.5035  & 0.4857  & 0.3592  & \textcolor[rgb]{ 1,  0,  0}{\textbf{0.8748}} & 0.7484  & \textcolor[rgb]{ 1,  0,  0}{\textbf{0.3879}} & 291.0118  \\
            & \checkmark     & \checkmark     & \textcolor[rgb]{ 1,  0,  0}{\textbf{0.4151}} & \textcolor[rgb]{ 1,  0,  0}{\textbf{0.8086}} & \textcolor[rgb]{ 1,  0,  0}{\textbf{0.5852}} & \textcolor[rgb]{ 1,  0,  0}{\textbf{0.5068}} & \textcolor[rgb]{ 1,  0,  0}{\textbf{0.3693}} & 0.8650  & \textcolor[rgb]{ 1,  0,  0}{\textbf{0.7792}} & 0.3714  & \textcolor[rgb]{ 1,  0,  0}{\textbf{283.3448}} \\
      \midrule
      \rowcolor[rgb]{ .859,  .859,  .859} Task  & TransBlock & 3 Transformations & $\rm{FMI}$ & $\rm{Q^{NICE}}$ & $\rm{Q^{M}}$ & $\rm{Q^{A/BF}}$ & $\rm{Q^P}$ & $\rm{SSIM}$ & $\rm{Q^Y}$ & $\rm{VIF}$ & $\rm{Q^{CV}}\downarrow$ \\
      \midrule
      \multirow{4}[2]{*}{\textit{MED}} &       &       & 0.2373  & 0.8087  & 0.1837  & 0.5254  & 0.4156  & 0.8891  & 0.4661  & 0.4998  & \textcolor[rgb]{ 1,  0,  0}{\textbf{860.2159}} \\
            & \checkmark     &       & 0.2373  & 0.8087  & 0.2049  & 0.3117  & 0.4500  & 0.8891  & 0.4865  & 0.4998  & 923.1668  \\
            &       & \checkmark    & 0.2838  & 0.8084  & 0.1971  & 0.5177  & 0.4491  & \textcolor[rgb]{ 1,  0,  0}{\textbf{0.9088}} & 0.4897  & \textcolor[rgb]{ 1,  0,  0}{\textbf{0.5347}} & 868.1148  \\
            & \checkmark     & \checkmark     & \textcolor[rgb]{ 1,  0,  0}{\textbf{0.3583}} & \textcolor[rgb]{ 1,  0,  0}{\textbf{0.8093}} & \textcolor[rgb]{ 1,  0,  0}{\textbf{0.2214}} & \textcolor[rgb]{ 1,  0,  0}{\textbf{0.5280}} & \textcolor[rgb]{ 1,  0,  0}{\textbf{0.4665}} & 0.9009  & \textcolor[rgb]{ 1,  0,  0}{\textbf{0.4997}} & 0.5147  & 938.6183  \\
      \midrule
      \rowcolor[rgb]{ .859,  .859,  .859} Task  & TransBlock & 3 Transformations & $\rm{FMI}$ & $\rm{Q^{NICE}}$ & $\rm{Q^{M}}$ & $\rm{Q^{A/BF}}$ & $\rm{Q^P}$ & $\rm{SSIM}$ & $\rm{Q^Y}$ & $\rm{VIF}$ & $\rm{Q^{CV}}\downarrow$ \\
      \midrule
      \multirow{4}[2]{*}{\textit{ME}} &       &       & 0.3269  & 0.8178  & 0.4540  & 0.6446  & 0.6345  & 0.9538  & 0.6581  & 1.0803  & 192.0676  \\
            & \checkmark     &       & 0.3723  & 0.8185  & 0.5159  & 0.6886  & 0.6973  & 0.9628  & 0.7070  & 1.0048  & 185.4562  \\
            &       & \checkmark    & 0.3917  & 0.8201  & 0.5162  & 0.6911  & 0.7107  & \textcolor[rgb]{ 1,  0,  0}{\textbf{0.9606}} & 0.6962  & \textcolor[rgb]{ 1,  0,  0}{\textbf{1.0959}} & \textcolor[rgb]{ 1,  0,  0}{\textbf{177.6318}} \\
            & \checkmark     & \checkmark     & \textcolor[rgb]{ 1,  0,  0}{\textbf{0.5199}} & \textcolor[rgb]{ 1,  0,  0}{\textbf{0.8262}} & \textcolor[rgb]{ 1,  0,  0}{\textbf{0.6482}} & \textcolor[rgb]{ 1,  0,  0}{\textbf{0.7318}} & \textcolor[rgb]{ 1,  0,  0}{\textbf{0.7642}} & 0.9594  & \textcolor[rgb]{ 1,  0,  0}{\textbf{0.7155}} & 1.0930  & 191.1575  \\
      \midrule
      \rowcolor[rgb]{ .859,  .859,  .859} Task  & TransBlock & 3 Transformations & $\rm{FMI}$ & $\rm{Q^{NICE}}$ & $\rm{Q^{M}}$ & $\rm{Q^{A/BF}}$ & $\rm{Q^P}$ & $\rm{SSIM}$ & $\rm{Q^Y}$ & $\rm{VIF}$ & $\rm{Q^{CV}}\downarrow$ \\
      \midrule
      \multirow{4}[2]{*}{\textit{MF}} &       &       & 0.3656  & 0.8305  & 0.4660  & 0.6920  & 0.7464  & 0.9826  & 0.8482  & 0.9903  & 79.0899  \\
            & \checkmark     &       & 0.4063  & 0.8315  & 0.5017  & 0.7186  & 0.7950  & 0.9839  & 0.8689  & 0.9950  & 72.6904  \\
            &       & \checkmark     & 0.4241  & 0.8324  & 0.5167  & 0.7244  & 0.8032  & 0.9845  & 0.8797  & \textcolor[rgb]{ 1,  0,  0}{\textbf{1.0108}} & 68.7478  \\
            & \checkmark     & \checkmark     & \textcolor[rgb]{ 1,  0,  0}{\textbf{0.5478}} & \textcolor[rgb]{ 1,  0,  0}{\textbf{0.8361}} & \textcolor[rgb]{ 1,  0,  0}{\textbf{0.6391}} & \textcolor[rgb]{ 1,  0,  0}{\textbf{0.7530}} & \textcolor[rgb]{ 1,  0,  0}{\textbf{0.8357}} & \textcolor[rgb]{ 1,  0,  0}{\textbf{0.9875}} & \textcolor[rgb]{ 1,  0,  0}{\textbf{0.9031}} & 0.9928  & \textcolor[rgb]{ 1,  0,  0}{\textbf{58.7026}} \\
      \bottomrule
      \end{tabular}%
    \label{table6}%
    }
  \end{table}%

\begin{figure*}[htbp]
    \centering
    \includegraphics[scale=0.5]{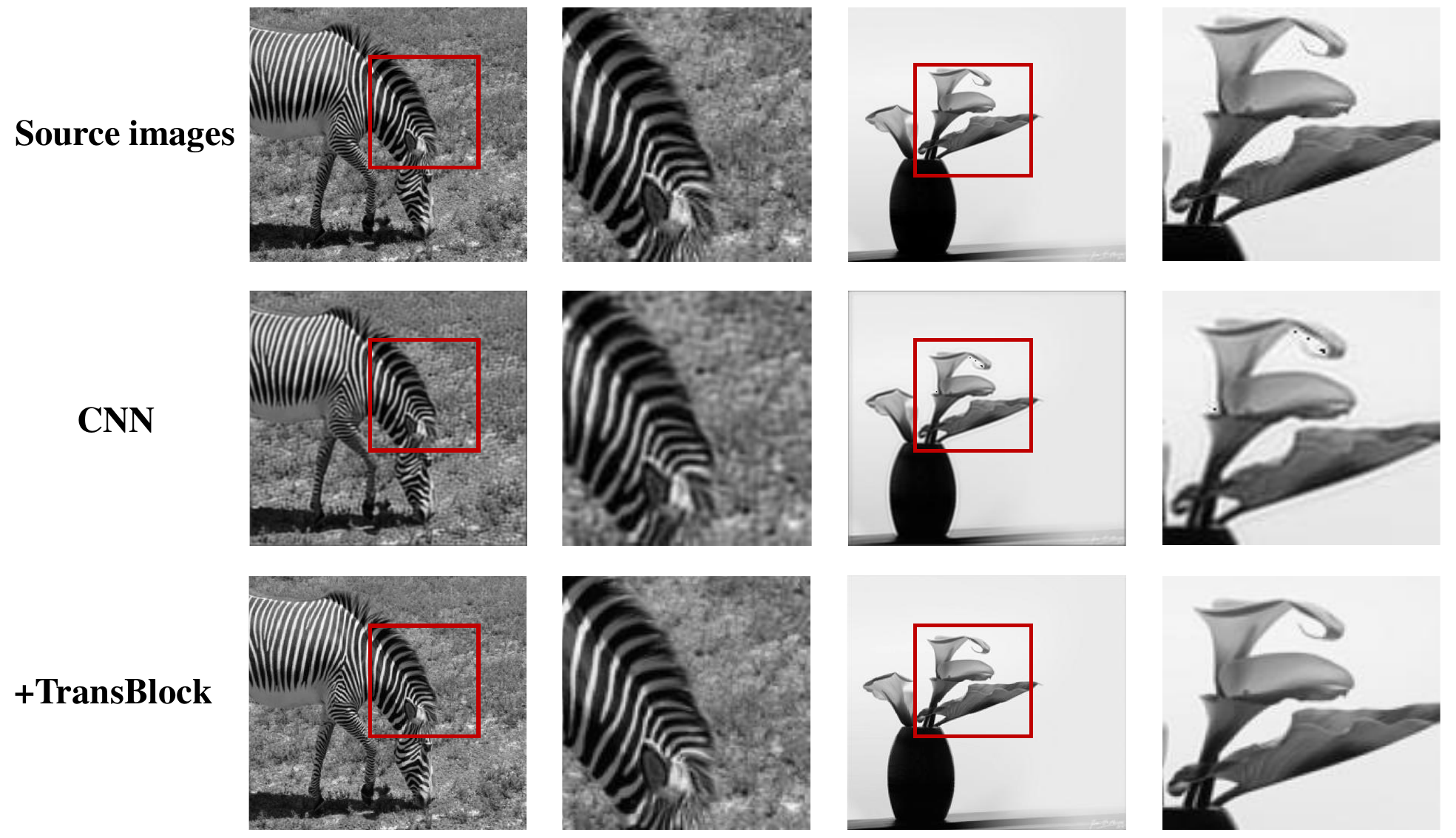}
    \caption{Visualization of the image reconstruction results using solely CNN and using TransBlock in the encoder. The top row is the original input image, the middle row is the image reconstruction results using traditional CNN architecture, and the bottom row is the image reconstruction results using TransBlock. The second and the fourth columns are the enlarged views of the red boxes shown in the first and the third columns, respectively.}
    \label{figure8}
\end{figure*}

To further explain the effectiveness of TransBlock, we visualize the image reconstruction results of using the CNN architecture (only the CNN-Module in TransBlock) and using TransBlock in Fig. \ref{figure8}. We can see that the reconstructed image with solely CNN lost some texture and detail information of the original image, such as the typical areas shown in the red box. In contrast, the reconstructed image with TransBlock has higher resolution and more detail information, which indicates that the encoding network with TransBlock performs better in extracting local and global features of the image.

\subsection{Ablation Study for Task-specific Self-Supervised Training Scheme}
We design task-specific self-supervised auxiliary tasks for multi-modal image fusion, multi-exposure image fusion, and multi-focus image fusion, which are respectively based on pixel intensity non-linear transformation, brightness transformation, and noise transformation. In this experiment, we evaluate the effectiveness of each auxiliary task and study if the performance is further improved by combing all the three transformation in our proposed model. The results are shown in Table \ref{table7}. Most of the fusion metrics are improved by using the corresponding self-supervised auxiliary tasks, which indicates the effectiveness of every single one of them. The effectiveness of our proposed combination strategy is demonstrated by the fact that most of the objective fusion metrics can be further improved by the proposed model.

\begin{table}[htbp]
    \centering
    \caption{Results of ablation tests on the auxiliary tasks using 20\% training data}
    \scalebox{0.55}{
      \begin{tabular}{ccccccccccccc}
      \toprule
      \rowcolor[rgb]{ .859,  .859,  .859} Task  & Nonlinear & Brightness & Noise & $\rm{FMI}$  & $\rm{Q^{NICE}}$ & $\rm{Q^{M}}$ & $\rm{Q^{A/BF}}$ & $\rm{Q^P}$ & $\rm{SSIM}$ & $\rm{Q^Y}$ & $\rm{VIF}$ & $\rm{Q^{CV}}\downarrow$ \\
      \midrule
      \multirow{3}[2]{*}{\textit{IV}} &       &       &       & 0.3135  & 0.8085  & 0.5172  & 0.4993  & 0.3606  & 0.8298  & 0.7773  & 0.3088  & 301.0812  \\
            & \checkmark     &       &       & 0.3522  & \textcolor[rgb]{ 1,  0,  0}{\textbf{0.8086}} & 0.5404  & 0.5061  & 0.3674  & 0.8489  & 0.7773  & 0.3451  & 296.2987  \\
            & \checkmark     & \checkmark     & \checkmark     & \textcolor[rgb]{ 1,  0,  0}{\textbf{0.4151}} & \textcolor[rgb]{ 1,  0,  0}{\textbf{0.8086}} & \textcolor[rgb]{ 1,  0,  0}{\textbf{0.5852}} & \textcolor[rgb]{ 1,  0,  0}{\textbf{0.5068}} & \textcolor[rgb]{ 1,  0,  0}{\textbf{0.3693}} & \textcolor[rgb]{ 1,  0,  0}{\textbf{0.8650}} & \textcolor[rgb]{ 1,  0,  0}{\textbf{0.7792}} & \textcolor[rgb]{ 1,  0,  0}{\textbf{0.3714}} & \textcolor[rgb]{ 1,  0,  0}{\textbf{283.3448}} \\
      \midrule
      \rowcolor[rgb]{ .859,  .859,  .859} Task  & Nonlinear & Brightness & Noise & $\rm{FMI}$  & $\rm{Q^{NICE}}$ & $\rm{Q^{M}}$ & $\rm{Q^{A/BF}}$ & $\rm{Q^P}$ & $\rm{SSIM}$ & $\rm{Q^Y}$ & $\rm{VIF}$ & $\rm{Q^{CV}}\downarrow$ \\
      \midrule
      \multirow{3}[2]{*}{\textit{MED}} &       &       &       & 0.2373  & 0.8087  & 0.2049  & 0.3117  & 0.4500  & 0.8891  & 0.4865  & 0.4998  & 923.1668  \\
            & \checkmark     &       &       & 0.2810  & 0.8088  & 0.2203  & \textcolor[rgb]{ 1,  0,  0}{\textbf{0.5329}} & 0.4653  & 0.8985  & 0.4961  & \textcolor[rgb]{ 1,  0,  0}{\textbf{0.5187}} & \textcolor[rgb]{ 1,  0,  0}{\textbf{915.8077}} \\
            & \checkmark     & \checkmark     & \checkmark     & \textcolor[rgb]{ 1,  0,  0}{\textbf{0.3583}} & \textcolor[rgb]{ 1,  0,  0}{\textbf{0.8093}} & \textcolor[rgb]{ 1,  0,  0}{\textbf{0.2214}} & 0.5280  & \textcolor[rgb]{ 1,  0,  0}{\textbf{0.4665}} & \textcolor[rgb]{ 1,  0,  0}{\textbf{0.9009}} & \textcolor[rgb]{ 1,  0,  0}{\textbf{0.4997}} & 0.5147  & 938.6183  \\
      \midrule
      \rowcolor[rgb]{ .859,  .859,  .859} Task  & Nonlinear & Brightness & Noise & $\rm{FMI}$  & $\rm{Q^{NICE}}$ & $\rm{Q^{M}}$ & $\rm{Q^{A/BF}}$ & $\rm{Q^P}$ & $\rm{SSIM}$ & $\rm{Q^Y}$ & $\rm{VIF}$ & $\rm{Q^{CV}}\downarrow$ \\
      \midrule
      \multirow{3}[2]{*}{\textit{ME}} &       &       &       & 0.3723  & 0.8185  & 0.5159  & 0.6886  & 0.6973  & 0.9628  & 0.7070  & 1.0048  & 185.4562  \\
            &       & \checkmark     &       & 0.4141  & 0.8208  & 0.5406  & 0.7084  & 0.7298  & \textcolor[rgb]{ 1,  0,  0}{\textbf{0.9598}} & 0.6991  & 1.0903  & \textcolor[rgb]{ 1,  0,  0}{\textbf{184.2343}} \\
            & \checkmark     & \checkmark     & \checkmark     & \textcolor[rgb]{ 1,  0,  0}{\textbf{0.5199}} & \textcolor[rgb]{ 1,  0,  0}{\textbf{0.8262}} & \textcolor[rgb]{ 1,  0,  0}{\textbf{0.6482}} & \textcolor[rgb]{ 1,  0,  0}{\textbf{0.7318}} & \textcolor[rgb]{ 1,  0,  0}{\textbf{0.7642}} & 0.9594  & \textcolor[rgb]{ 1,  0,  0}{\textbf{0.7155}} & \textcolor[rgb]{ 1,  0,  0}{\textbf{1.0930}} & 191.1575  \\
      \midrule
      \rowcolor[rgb]{ .859,  .859,  .859} Task  & Nonlinear & Brightness & Noise & $\rm{FMI}$  & $\rm{Q^{NICE}}$ & $\rm{Q^{M}}$ & $\rm{Q^{A/BF}}$ & $\rm{Q^P}$ & $\rm{SSIM}$ & $\rm{Q^Y}$ & $\rm{VIF}$ & $\rm{Q^{CV}}\downarrow$ \\
      \midrule
      \multirow{3}[2]{*}{\textit{MF}} &       &       &       & 0.4063  & 0.8315  & 0.5017  & 0.7186  & 0.7950  & 0.9839  & 0.8689  & 0.9950  & 72.6904  \\
            &       &       & \checkmark     & 0.4561  & 0.8334  & 0.5539  & 0.7387  & 0.8235  & 0.9857  & 0.8904  & \textcolor[rgb]{ 1,  0,  0}{\textbf{1.0081}} & 64.0137  \\
            & \checkmark     & \checkmark     & \checkmark     & \textcolor[rgb]{ 1,  0,  0}{\textbf{0.5478}} & \textcolor[rgb]{ 1,  0,  0}{\textbf{0.8361}} & \textcolor[rgb]{ 1,  0,  0}{\textbf{0.6391}} & \textcolor[rgb]{ 1,  0,  0}{\textbf{0.7530}} & \textcolor[rgb]{ 1,  0,  0}{\textbf{0.8357}} & \textcolor[rgb]{ 1,  0,  0}{\textbf{0.9875}} & \textcolor[rgb]{ 1,  0,  0}{\textbf{0.9031}} & 0.9928  & \textcolor[rgb]{ 1,  0,  0}{\textbf{58.7026}} \\
      \bottomrule
      \end{tabular}%
    \label{table7}%
    }
  \end{table}%

  \begin{table}[htbp]
    \centering
    \caption{Results of ablation tests on the number and size of the transformed subregions.}
    \scalebox{0.5}{
      \begin{tabular}{ccccccccccc}
      \toprule
      \rowcolor[rgb]{ .859,  .859,  .859} Task  & Method & $\rm{FMI}$  & $\rm{Q^{NICE}}$ & $\rm{Q^{M}}$ & $\rm{Q^{A/BF}}$ & $\rm{Q^P}$ & $\rm{SSIM}$ & $\rm{Q^Y}$ & $\rm{VIF}$ & $\rm{Q^{CV}}\downarrow$ \\
      \midrule
      \multirow{13}[2]{*}{\textit{IV}} & N1\_R256 & 0.1947  & 0.8054  & 0.3667  & 0.3672  & 0.1800  & 0.8288  & 0.3382  & 0.6367  & 392.0450  \\
            & N2\_R8 & 0.3307  & 0.8087  & 0.5250  & 0.5019  & 0.3639  & 0.8455  & 0.3423  & 0.7716  & 297.5007  \\
            & N2\_R16 & 0.3319  & 0.8085  & \textcolor[rgb]{ 1,  0,  0}{\textbf{0.5379}} & \textcolor[rgb]{ 1,  0,  0}{\textbf{0.5053}} & 0.3625  & 0.8525  & 0.3253  & 0.7667  & 300.5854  \\
            & N2\_R32 & 0.3282  & 0.8085  & 0.5258  & 0.5005  & 0.3618  & 0.8486  & 0.3363  & 0.7826  & 300.3576  \\
            & N2\_R64 & 0.3122  & 0.8081  & 0.5010  & 0.4952  & 0.3551  & 0.8414  & 0.3214  & 0.7747  & 298.4003  \\
            & N4\_R8 & 0.3163  & 0.8083  & 0.5222  & 0.5007  & 0.3563  & 0.8524  & 0.3331  & 0.7764  & 301.0319  \\
            & N4\_R16 & \textcolor[rgb]{ 1,  0,  0}{\textbf{0.3405}} & \textcolor[rgb]{ 1,  0,  0}{\textbf{0.8088}} & \textcolor[rgb]{ 1,  0,  0}{\textbf{0.5379}} & 0.5035  & \textcolor[rgb]{ 1,  0,  0}{\textbf{0.3642}} & 0.8367  & 0.3270  & \textcolor[rgb]{ 1,  0,  0}{\textbf{0.7872}} & 303.8792  \\
            & N4\_R32 & 0.3169  & 0.8083  & 0.5162  & 0.4973  & 0.3538  & 0.8467  & 0.3328  & 0.7821  & 300.4295  \\
            & N4\_R64 & 0.2702  & 0.8060  & 0.4390  & 0.4528  & 0.3021  & 0.8646  & \textcolor[rgb]{ 1,  0,  0}{\textbf{0.3448}} & 0.7247  & 341.9801  \\
            & N8\_R8 & 0.2702  & 0.8060  & 0.5336  & 0.4528  & 0.3613  & \textcolor[rgb]{ 1,  0,  0}{\textbf{0.8646}} & \textcolor[rgb]{ 1,  0,  0}{\textbf{0.3448}} & 0.7811  & 295.8375  \\
            & N8\_R16 & 0.3272  & 0.8086  & 0.5244  & 0.5034  & 0.3637  & 0.8480  & 0.3446  & 0.7776  & \textcolor[rgb]{ 1,  0,  0}{\textbf{295.2036}} \\
            & N8\_R32 & 0.2964  & 0.8075  & 0.4949  & 0.4826  & 0.3348  & 0.8565  & 0.3246  & 0.7694  & 315.0558  \\
            & N8\_R64 & 0.2575  & 0.8053  & 0.4228  & 0.4184  & 0.2559  & 0.8211  & 0.2717  & 0.6960  & 428.7837  \\
      \midrule
      \multirow{13}[2]{*}{\textit{MED}} & N1\_R256 & 0.2075  & 0.8105  & 0.3116  & 0.3584  & 0.2832  & \textcolor[rgb]{ 1,  0,  0}{\textbf{0.8848}} & 0.8795  & 0.4563  & 306.8308  \\
            & N2\_R8 & 0.3695  & 0.8155  & 0.4446  & 0.6374  & 0.5436  & 0.8341  & 1.1028  & 0.7219  & 214.4595  \\
            & N2\_R16 & 0.3753  & 0.8158  & 0.4490  & \textcolor[rgb]{ 1,  0,  0}{\textbf{0.6450}} & 0.5548  & 0.8380  & 1.0856  & \textcolor[rgb]{ 1,  0,  0}{\textbf{0.7313}} & 219.3057  \\
            & N2\_R32 & 0.3663  & 0.8152  & 0.4321  & 0.6313  & 0.5482  & 0.8334  & 1.0987  & 0.7141  & \textcolor[rgb]{ 1,  0,  0}{\textbf{211.9355}} \\
            & N2\_R64 & 0.3365  & 0.8121  & 0.4374  & 0.6065  & 0.4909  & 0.8277  & 0.9498  & 0.6878  & 305.5588  \\
            & N4\_R8 & 0.3536  & 0.8147  & 0.4365  & 0.6300  & 0.5440  & 0.8392  & 1.0878  & 0.7145  & 219.4811  \\
            & N4\_R16 & \textcolor[rgb]{ 1,  0,  0}{\textbf{0.3823}} & \textcolor[rgb]{ 1,  0,  0}{\textbf{0.8159}} & \textcolor[rgb]{ 1,  0,  0}{\textbf{0.4556}} & 0.6402  & 0.5491  & 0.8272  & \textcolor[rgb]{ 1,  0,  0}{\textbf{1.1091}} & 0.7224  & 224.6704  \\
            & N4\_R32 & 0.3503  & 0.8144  & 0.4311  & 0.6209  & 0.5300  & 0.8366  & 1.0330  & 0.7083  & 220.9103  \\
            & N4\_R64 & 0.2873  & 0.8091  & 0.4061  & 0.5632  & 0.3881  & 0.8151  & 0.8239  & 0.6460  & 259.3240  \\
            & N8\_R8 & 0.2873  & 0.8091  & 0.4363  & 0.5632  & \textcolor[rgb]{ 1,  0,  0}{\textbf{0.5573}} & 0.8151  & 0.8239  & 0.7218  & 212.2687  \\
            & N8\_R16 & 0.3646  & 0.8151  & 0.4336  & 0.6261  & 0.5449  & 0.8323  & 1.1038  & 0.7199  & 214.0531  \\
            & N8\_R32 & 0.3195  & 0.8115  & 0.4452  & 0.5948  & 0.4620  & 0.8441  & 0.8939  & 0.6812  & 325.3530  \\
            & N8\_R64 & 0.2722  & 0.8087  & 0.4074  & 0.5372  & 0.3868  & 0.8361  & 0.6873  & 0.6441  & 435.6439  \\
      \midrule
      \multirow{13}[2]{*}{\textit{ME}} & N1\_R256 & 0.1518  & 0.8061  & 0.1305  & 0.2886  & 0.1937  & 0.8387  & 0.4201  & 0.3354  & 1525.4510  \\
            & N2\_R8 & 0.2634  & 0.8077  & \textcolor[rgb]{ 1,  0,  0}{\textbf{0.1568}} & \textcolor[rgb]{ 1,  0,  0}{\textbf{0.4223}} & \textcolor[rgb]{ 1,  0,  0}{\textbf{0.3880}} & 0.8930  & 0.4913  & \textcolor[rgb]{ 1,  0,  0}{\textbf{0.4365}} & 1171.2512  \\
            & N2\_R16 & 0.2636  & 0.8078  & 0.1553  & 0.4114  & 0.3853  & 0.8883  & 0.4844  & 0.4286  & 1186.1036  \\
            & N2\_R32 & 0.2501  & 0.8076  & 0.1536  & 0.4075  & 0.3814  & 0.8884  & 0.4848  & 0.4240  & 1199.2137  \\
            & N2\_R64 & 0.2349  & 0.8074  & 0.1520  & 0.4068  & 0.3688  & 0.8853  & 0.4868  & 0.4235  & 1218.6177  \\
            & N4\_R8 & 0.2562  & 0.8076  & 0.1532  & 0.4103  & 0.3833  & 0.8905  & 0.4857  & 0.4235  & \textcolor[rgb]{ 1,  0,  0}{\textbf{1161.3674}} \\
            & N4\_R16 & \textcolor[rgb]{ 1,  0,  0}{\textbf{0.2637}} & \textcolor[rgb]{ 1,  0,  0}{\textbf{0.8078}} & 0.1553  & 0.4214  & 0.3859  & \textcolor[rgb]{ 1,  0,  0}{\textbf{0.8983}} & \textcolor[rgb]{ 1,  0,  0}{\textbf{0.4944}} & 0.4286  & 1186.1036  \\
            & N4\_R32 & 0.2405  & 0.8075  & 0.1521  & 0.4061  & 0.3742  & 0.8900  & 0.4908  & 0.4299  & 1202.8124  \\
            & N4\_R64 & 0.2011  & 0.8068  & 0.1445  & 0.3632  & 0.3224  & 0.8573  & 0.4441  & 0.3914  & 1452.7672  \\
            & N8\_R8 & 0.2591  & 0.8077  & 0.1535  & 0.4113  & 0.3857  & 0.8908  & 0.4872  & 0.4296  & 1175.6538  \\
            & N8\_R16 & 0.2578  & 0.8076  & 0.1542  & 0.4103  & 0.3844  & 0.8926  & 0.4884  & 0.4279  & 1186.3845  \\
            & N8\_R32 & 0.2293  & 0.8072  & 0.1493  & 0.3753  & 0.3565  & 0.8745  & 0.4626  & 0.4047  & 1376.2388  \\
            & N8\_R64 & 0.1908  & 0.8063  & 0.1393  & 0.3339  & 0.2599  & 0.7804  & 0.3543  & 0.3720  & 2190.5549  \\
      \midrule
      \multirow{13}[2]{*}{\textit{MF}} & N1\_R256 & 0.2453  & 0.8219  & 0.3296  & 0.4923  & 0.4733  & 0.9256  & \textcolor[rgb]{ 1,  0,  0}{\textbf{1.0105}} & 0.6825  & 173.9629  \\
            & N2\_R8 & 0.4321  & 0.8328  & 0.5291  & 0.7311  & 0.8086  & 0.9801  & 0.9967  & 0.8823  & 72.3116  \\
            & N2\_R16 & 0.4354  & 0.8329  & 0.5324  & \textcolor[rgb]{ 1,  0,  0}{\textbf{0.7327}} & 0.8102  & 0.9845  & 0.9948  & \textcolor[rgb]{ 1,  0,  0}{\textbf{0.8850}} & \textcolor[rgb]{ 1,  0,  0}{\textbf{70.5472}} \\
            & N2\_R32 & 0.4279  & 0.8326  & 0.5208  & 0.7274  & 0.8058  & 0.9843  & 0.9984  & 0.8820  & 74.6641  \\
            & N2\_R64 & 0.4024  & 0.8292  & 0.4937  & 0.7146  & 0.7901  & 0.9833  & 0.9635  & 0.8657  & 81.7729  \\
            & N4\_R8 & 0.4152  & 0.8322  & 0.5117  & 0.7246  & 0.7993  & 0.9839  & 0.9956  & 0.8733  & 73.1741  \\
            & N4\_R16 & \textcolor[rgb]{ 1,  0,  0}{\textbf{0.4421}} & \textcolor[rgb]{ 1,  0,  0}{\textbf{0.8331}} & \textcolor[rgb]{ 1,  0,  0}{\textbf{0.5338}} & 0.7311  & \textcolor[rgb]{ 1,  0,  0}{\textbf{0.8103}} & \textcolor[rgb]{ 1,  0,  0}{\textbf{0.9850}} & 0.9825  & 0.8839  & 74.4669  \\
            & N4\_R32 & 0.4119  & 0.8315  & 0.5038  & 0.7214  & 0.7970  & 0.9839  & 1.0001  & 0.8751  & 76.0166  \\
            & N4\_R64 & 0.3498  & 0.8256  & 0.4420  & 0.6660  & 0.7205  & 0.9742  & 0.8940  & 0.8151  & 101.5177  \\
            & N8\_R8 & 0.3498  & 0.8256  & 0.5293  & 0.6660  & 0.8100  & 0.9742  & 0.8940  & 0.8819  & 72.5735  \\
            & N8\_R16 & 0.4290  & 0.8326  & 0.5216  & 0.7293  & 0.8087  & 0.9841  & 0.9999  & 0.8823  & 73.7230  \\
            & N8\_R32 & 0.3846  & 0.8295  & 0.4725  & 0.7023  & 0.7746  & 0.9817  & 0.9561  & 0.8510  & 86.1646  \\
            & N8\_R64 & 0.3297  & 0.8222  & 0.4106  & 0.6277  & 0.6543  & 0.9413  & 0.7608  & 0.7628  & 153.8210  \\
      \bottomrule
      \end{tabular}%
    \label{table8}%
    }
  \end{table}%
\subsection{Ablation Study for the Number and Area of Transformed Subregion}
In our proposed method, we randomly generate four image subregions with the size of 16$\times$16 from a 256$\times$256 training image to form the set $chi$ for transformation. Obviously, both the number and size of transform subregions will affect the efficiency of trained network. For example, when the size of the subregion is too large (even transforming the whole image), it will be very difficult for the network to reconstruct the original image. At the same time, because most of the original image is destroyed, the network cannot learn useful features. On the contrary, when the subregions are too small, the effect of the transformation will also be small and may not be able to encourage the network to learn better task-specific features. We conducted the ablation experiments on different combinations of the number and size of the subregions, and the results are shown in Table \ref{table8}. In this experiment, we used 20\% of the training data with 30 epochs.

In Table \ref{table8}, $N$ represents the number of transformed regions and $R$ represents the size of transformed subregions. For example, $N1\_R256$ represents the transformation of the whole image, and $N4\_R16$ represents the transformation of four randomly selected 16$\times$16 regions. It can be seen that $N4\_R16$ achieves the best performance overall.

\section{Discussion}
In this study, we propose TransFuse, a unified Transformer-based image fusion framework by self-supervised learning, which can be effectively used to different image fusion tasks, including multi-modal, multi-exposure and multi-focus image fusion. We propose three destruction-reconstruction self-supervised auxiliary tasks for multi-modal, multi-exposure and multi-focus image fusion and integrate the three tasks by randomly choosing one of them to destroy a natural image during model training, thus enabling our network to be trained on accessible large natural image datasets and can learn task-specific features at the same time. In addition, we design a new encoder that combines CNN and Transformer for feature extraction, so that the model can exploit both local and global information more comprehensively. Extensive experiments have shown that our framework achieves new state-of-the-art performance in both subjective and objective evaluations in all common image fusion tasks.

Notably, in different fusion tasks, the vital information to be fused varies largely as source images contain different characteristics. The three destruction-reconstruction self-supervised auxiliary tasks are specially designed according to the characteristics of source images in different fusion tasks, so that our framework can learn task-specific features during reconstruction. Furthermore, we integrate the three tasks in model training to encourage different fusion tasks to promote each other and increase the generalization of the trained network, enabling our framework to handle different image fusion tasks in a unified way. 

Our framework also has some limitations. First, although we utilize parameter-shared Fine-grained Transformers, introducing Transformer for feature extraction still makes our model (with 43.18 MB parameters) larger than the existing methods (generally with 0.3-3 MB parameters). Fortunately, a common GPU NVIDIA 1080Ti is still sufficient for model training. Second, we do not propose new fusion rules for different fusion tasks and more effective task-specific fusion rules can be specially designed in the future to further improve the fusion performance.

In future work, more effective Transformer-based feature extraction methods can be explored for image fusion tasks. Moreover, since both CNN-based architectures and Transformer-based architectures have their advantages, how to further combine the two architectures is another promising research direction. On the other hand, more effective self-supervised auxiliary tasks may also be proposed to enforce the network learn more useful features according to the characteristics of different source images. 

In addition, our proposed self-supervised tasks and training scheme can help the model learn more robust and generalized features, and our proposed Transformer-based feature extraction modules can exploit both local and global information more comprehensively, hence, we expect that our method has a broad application prospect and a large development potential in both image fusion field and related computer vision fields.

\bibliography{mybibfile}

\end{document}